\documentclass[10pt,twocolumn,letterpaper]{article}

\usepackage{cvpr}
\makeatletter
\@namedef{ver@everyshi.sty}{}
\makeatother
\usepackage[pdftex,dvipsnames]{xcolor}  
\usepackage{xargs}                      
\usepackage{todonotes}      
\newcommandx{\vivek}[1]{\todo[inline,linecolor=Blue,backgroundcolor=BlueGreen,bordercolor=Blue]{\small\textsf{\textbf{Vivek}: #1}}}
\newcommandx{\aaron}[1]{\todo[inline,linecolor=RawSienna,backgroundcolor=Tan,bordercolor=RawSienna]{\small\textsf{\textbf{Aaron}: #1}}}
\newcommandx{\basil}[1]{\todo[inline,linecolor=Plum,backgroundcolor=Plum!25,bordercolor=Plum]{\small\textsf{\textbf{Basil}: #1}}}
\newcommandx{\neil}[1]{\todo[inline,linecolor=OliveGreen,backgroundcolor=OliveGreen!25,bordercolor=OliveGreen]{\small\textsf{\textbf{Neil}: #1}}}
\newcommandx{\jan}[1]{\todo[inline,linecolor=orange,backgroundcolor=orange!35,bordercolor=orange]{\small\textsf{\textbf{Jan}: #1}}}

\usepackage{enumitem} 
\usepackage[skip=1pt]{caption}
\usepackage{subcaption}
\usepackage{floatrow}
\usepackage{booktabs, multirow}
\definecolor{darkgray}{rgb}{0.25, 0.25, 0.25}
\newcommand{\rng}[2]{\textcolor{darkgray}{\textsubscript{[#1 - #2]}}}

\usepackage{selectp} 
\usepackage{times}
\usepackage{epsfig}
\usepackage{graphicx}
\usepackage{amsmath}
\usepackage{amssymb}

\usepackage[para]{footmisc}%

\usepackage[breaklinks=true,bookmarks=false]{hyperref}

\cvprfinalcopy 

\def\cvprPaperID{10141} 

\renewcommand{\paragraph}[1]{{\bf #1}\,\,}

\begin{document}

\title{Supervised Transfer Learning at Scale for Medical Imaging}

\author{
Basil Mustafa $\cdot$ 
Aaron Loh  $\cdot$ Jan Freyberg $\cdot$ Patricia MacWilliams \\
Megan Wilson $\cdot$ Scott Mayer McKinney $\cdot$ Marcin Sieniek  $\cdot$ Jim Winkens \\
Yuan Liu $\cdot$ Peggy Bui $\cdot$ Shruthi Prabhakara $\cdot$ Umesh Telang \\
Alan Karthikesalingam $\cdot$ Neil Houlsby $\cdot$ Vivek Natarajan \\
Google Research \& Google Health\\
}

\maketitle
\let\thefootnote\relax\footnotetext{Correspondence:\\\tt\{basilm,janfreyberg,neilhoulsby,natviv\}@google.com}

\begin{abstract}
Transfer learning is a standard technique to improve performance on tasks with limited data.
However, for medical imaging, the value of transfer learning is less clear~\cite{raghu2019transfusion}.
This is likely due to the large domain mismatch between the usual natural-image pre-training (e.g. ImageNet) and medical images.
However, recent advances in transfer learning have shown substantial improvements from scale.
We investigate whether modern methods can change the fortune of transfer learning for medical imaging.
For this, we study the class of large-scale pre-trained networks presented by Kolesnikov \textit{et al.}~\cite{kolesnikov2019big} on three diverse imaging tasks: chest radiography, mammography, and dermatology.
We study both transfer performance and critical properties for the deployment in the medical domain, including: out-of-distribution generalization, data-efficiency, sub-group fairness, and uncertainty estimation.
Interestingly, we find that for some of these properties, transfer from natural to medical images is indeed extremely effective, but only when performed at sufficient scale.
\end{abstract}

\section{Introduction}

Deep learning has enabled many exciting recent advances to medical imaging.
High performing models have been developed, often competitive with human experts, in a variety of domains including ophthalmology~\cite{gulshan2016development}, radiology~\cite{mckinney2020international, rajpurkar2017chexnet}, dermatology~\cite{esteva2017dermatologist, liu2020deep} and pathology~\cite{liu2017detecting}. 

A key component of many successful models is transfer learning, where models are pre-trained on a source dataset, and fine-tuned on the target task.
This is particularly effective when target data is limited or expensive, as is often the case in medical imaging.
ILSVRC-2012 (ImageNet)~\cite{imagenet_cvpr09} is the most popular pre-training dataset, and pre-training on it improves performance across wide range of computer vision tasks, including classification, detection, and segmentation~\cite{yosinski2014transferable,huh2016makes,hermans2017,long2015fully, girshick2014}.

However, the effectiveness of transfer learning is mostly validated on natural image datasets, which form standard computer vision benchmarks.
Recent works analyzing the role of transfer learning for medical imaging suggest that its benefits may be limited~\cite{raghu2019transfusion,alzubaidi2020towards} that pre-training on in-domain medical imaging data is more effective~\cite{alzubaidi2020towards}. 
This is not surprising; medical images present a large domain shift when transferring from natural images, such as those in ImageNet. 
The images are typically much higher resolution, with non-RGB channels, and models often rely on small, local variations in texture to interpret the images and detect pathologies of interest. 

That being said, recent works demonstrate large improvements in transfer performance over ImageNet pre-training~\cite{mahajan2018exploring,kolesnikov2019big,xie2020noisystudent}.
These papers show that a key component of the efficacy of transfer learning is \textit{scale}, both in terms of the model capacity and the size of the pre-training dataset. 
However, they still predominantly focus on natural-world images, with massive pre-training datasets typically scraped from social media or the Web and evaluations performed on popular datasets such as ImageNet or CIFAR~\cite{cifar100}.
There is, however, some indication that such pre-training at scale on natural datasets can help more diverse tasks~\cite{zhai2019large}.
Here, we seek to understand whether modern large-scale pre-training is effective for medical imaging.

\begin{figure*}[htbp]
\floatbox[{\capbeside\thisfloatsetup{capbesideposition={right,top},capbesidewidth=0.3\textwidth}}]{figure}[\FBwidth]
{\caption{
Transfer learning is well established for natural image tasks, and ImageNet is frequently used for pre-training and/or evaluation.
Further, like-to-like transfer within the medical domain has been shown to work~\cite{heker2020joint, liang2020transfer, pmlr-v102-geyer19a, chen2019med3d}.
However, the effectiveness of transfer from natural image datasets to medical imaging is debated~\cite{raghu2019transfusion}.
We study this regime in the context of modern transfer methods to better understand the state of the field in this important, yet challenging, domain.}\label{fig:collage}}
{\includegraphics[width=0.65\textwidth]{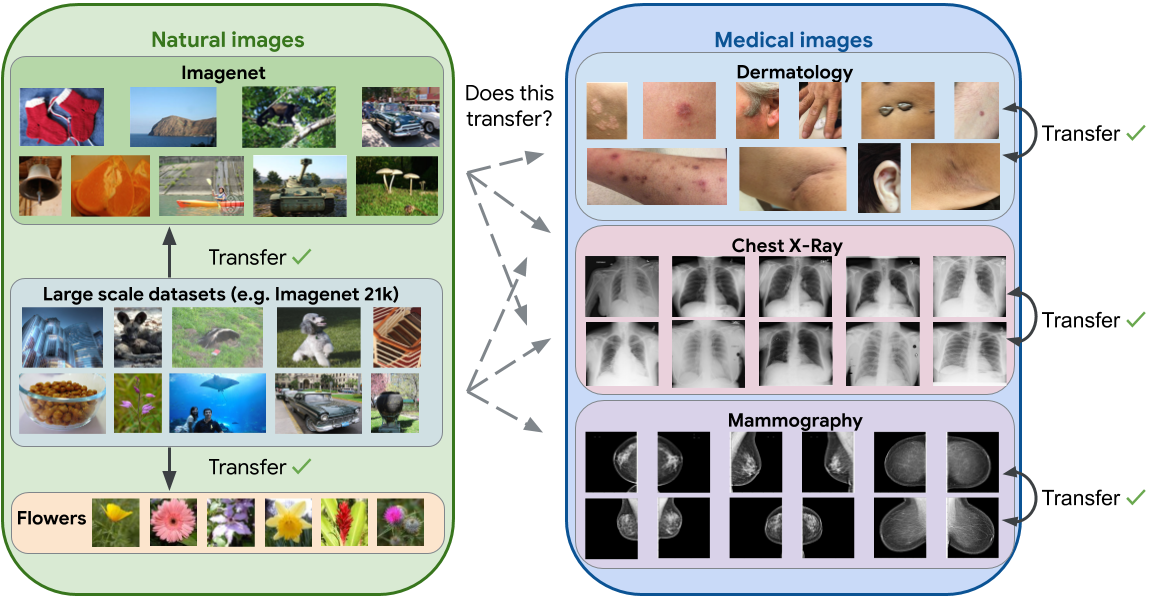}}
\end{figure*}

We focus on the class of models presented in~\cite{kolesnikov2019big}. 
These models, termed ``Big Transfer'' (BiT), attained state of the art on many visual classification benchmarks. 
These models follow ResNet architectures~\cite{he2015resnetv2} with small architectural tweaks which optimise transfer performance.
Importantly for our study, they span the main dimensions of modern effective transfer  learning:
they are pre-trained on different datasets from 1M to 300M images, and they range in architecture size from 25M parameters up to nearly 1B.
We study these key components in depth on three medical imaging tasks: mammography cancer identification, chest x-ray interpretation, and skin condition classification.
The tasks vary in characteristics, and differ greatly from the usual success cases demonstrated by large scale transfer learning.

To understand the implication of transfer of medical imaging, we not only look at standard performance metrics, but also include analysis of key properties particularly important to training and deploying models. Our main contributions are summarized as follows.

\begin{itemize}[itemsep=1pt,topsep=1pt,leftmargin=12pt]
\item We demonstrate that supervised transfer learning at scale can indeed lead to superior performance on medical imaging tasks.  These results are significant especially in light of prior works suggesting benefits of such transfer learning for medical tasks may be limited \cite{raghu2019transfusion, neyshabur2020being}.
\item We find that such large scale pre-trained models yield improved generalization under distribution shifts and data efficiency. For example, the models are able to match the baseline performance using between 30-60\% of the training data.
\item We find that these improvements do not come at the expense of subgroup fairness or model calibration; which are important considerations for model deployment in clinical settings.
\item Finally, to better understand the source of these improvements, we conduct analysis of the model weights and features. Our exploratory results suggest that pretraining at scale may allow downstream tasks to better reuse deeper feature more effectively, something previous \cite{raghu2019transfusion} studies indicate is less useful. More comprehensive study is warranted here.
\end{itemize}



\section{Related Work}

\subsection{Transfer learning for medical imaging}

Transfer learning is a commonly adopted strategy when building medical imaging models. The reasoning behind adopting this strategy is the often limited availability of medical imaging data\footnote{Willemink \textit{et al.}~
\cite{willemink2020preparing} summarize the challenges involved in the curation of medical imaging datasets for machine learning.} which may prevent learning of good representations when models are trained from scratch without pre-training. Typically, standard architectures pre-trained on ImageNet are chosen as the starting point to fine-tune models on downstream medical classification tasks (e.g.~\cite{xie2019dual, menegola2017knowledge, mckinney2020international, liu2020deep, he2020sample}).\\
However, the natural images in datasets like ImageNet significantly differ from typical medical images as can be seen in Figure \ref{fig:collage}. Often, medical images for a given domain have standardized views; images are frequently gray-scale; resolution is usually significantly higher; and the task relevant features tend to be small patches or local texture variations instead of semantic higher level features. 
In some cases, detailed experiments have been performed to demonstrate the benefits of adopting this transfer~\cite{alzubaidi2020towards, graziani2019visualizing, heker2020joint}, but it is not clear if this is always beneficial.

Though previous works~\cite{kornblith2019better} suggest better ImageNet classification performance leads to better downstream performance on natural image tasks, this may not necessarily hold for medical imaging tasks.
Towards better understanding the benefits and trade offs of ImageNet based transfer learning for medical imaging, Raghu \textit{et al.}~\cite{raghu2019transfusion} perform detailed experiments on two large scale medical datasets, chest X-Rays (CheXpert~\cite{irvin2019chexpert}), and retinal fundus images~\cite{gulshan2016development}. Their experiments suggest that the domain mismatch between natural and medical images inhibits transfer learning; across a range of architectures, performance did not increase significantly when initialized with ImageNet pre-trained weights. Moreover, using carefully crafted, randomly initialized smaller architectures, the authors are able to achieve similar task performance as larger ImageNet pre-trained models. These results are in line with Neyshabur \textit{et al.}~\cite{neyshabur2020being} whose experiments suggest the key contributions to performance gains in transfer learning come from learning low-level image statistics and feature reuse. Both aspects differ between ImageNet and medical images such as X-Rays.

Despite this headline result, Raghu \textit{et al.} observe two clear benefits of transfer learning. First, when simulating a low-data regime, they find their largest model improves when pre-trained. Second, pre-training significantly speeds up convergence. This correlates with our findings here, although it is worth pointing out that the largest model they consider is equivalent in capacity to the \textit{smallest} we study.

\subsection{Large Scale Pre-training}

The ILSVRC-2012 ImageNet challenge dataset is the most popular for pre-training, containing 1.3M images~\cite{russakovsky2015imagenet}.
Recently, much larger datasets have been explored.
Early work using datasets much larger than ImageNet trained on 100M Flickr images, labelled with text captions~\cite{joulin2016learning,li2017learning}.
This has been extended much further, and a dataset of up to 3.5B images from Instagram weakly-labelled with hashtags has been explored, attaining strong results in transfer learning~\cite{mahajan2018exploring}.
Billion-scale datasets from social media have also been used for semi-supervised learning~\cite{yalniz2019billion}.
Another large dataset is JFT-300M, which consists of 300M labelled images from the web~\cite{jft300m}.
The labels are noisy, with a 20\% reported precision errors~\cite{jft300m}.
Recently, very strong image classification results have been attained using JFT-300M as auxiliary data~\cite{kolesnikov2019big,noisystudent,dosovitskiy2020}.
\cite{jft300m,kolesnikov2019big} train using the 18k JFT labels, and~\cite{noisystudent} performs semi-supervised learning using JFT-300M images and labelled ImageNet.


While there has been work exploring transfer from ImageNet, or smaller in-domain datasets to medical imaging tasks, there is not systematic evaluation of recent large-scale pre-training methods to the best of our knowledge.


\subsection{In-domain pre-training}

While we largely discuss transfer learning from natural to medical images, it is worth noting that recent works in medical imaging~\cite{heker2020joint, liang2020transfer, pmlr-v102-geyer19a, chen2019med3d} suggest in-domain pre-training is also very effective. For example,~\cite{heker2020joint} compares using weights initialized from an ImageNet-trained model to using weights trained on an in-domain dataset for the purposes of liver segmentation. They find pre-training on a separate liver imaging dataset improves performance more than ImageNet pre-training.

However, as medical data is often expensive and time-consuming to annotate, curating large, labeled in-domain datasets for pre-training is a challenge. Advances in self-supervised learning~\cite{chen2020simple, he2020sample} may enable training on unlabelled medical datasets, but even those are often hard to come by. As advances there may be complementary to out-of-domain pre-training, we believe that effective large scale transfer from natural images still has a valuable role to play in building performant medical imaging models regardless of -- or perhaps in conjunction with -- other advances.

\section{Transfer Learning at Scale for Medical Imaging}

Transfer from ImageNet pre-trained models is still the norm.
However, as discussed above, significant recent advances have appeared in large-scale pre-training.
By studying a range of larger pre-trained models on a diverse selection of downstream medical tasks, we aim to establish whether the benefits from recent advances transfer to medical imaging.

Our models consist predominately of a Convolutional Neural Network (CNN) ``backbone'', which is pre-trained.
Each task then has specific modelling features such as prediction heads and custom losses. 
For example, the chest X-Ray model follows a standard paradigm of predicting from features extracted from a single image, but the mammography model concatenates features extracted from four different images, and the dermatology model averages features from a variable number (1-6) of images.
We note that these models have significantly distinct setups, however, these setups are necessary to be able to draw meaningful conclusions at state-of-the-art quality levels~\cite{mckinney2020international,liu2017detecting}. 
Full details of the task-specific setups are in Appendix~\ref{app:modelling}.

For transfer we follow the usual paradigm; model weights are pre-trained on a \emph{source} or \emph{upstream} task. The CNN backbone is initialized to these pre-trained weights, and then subsequently fine-tuned\footnote{While other adaptation methods exist (e.g.~\cite{rebuffi2017}), fine-tuning tends to perform best~\cite{kornblith2019better,zhai2019large}.} alongside task-specific components on the \emph{target} or \emph{downstream} task.

\subsection{Methods}

\paragraph{Baseline}
We compare the effects of modern supervised pre-training against an industry and research standard; a ResNet50-v2 (R50x1) with Batch Normalization pre-trained on ILSVRC2012 ImageNet~\cite{tfslim_baseline_model}. This model is common for vision tasks, including medical imaging, and is used in previous medical imaging transfer learning studies~\cite{raghu2019transfusion,neyshabur2020being} and the previous Mammography study \cite{mckinney2020international}.

\paragraph{Big Transfer Models}
The BiT models are based on varying sizes ResNet-v2\cite{he2015resnetv2} architectures, pre-trained with supervised learning on natural datasets of different scales.

The architecture is the same as the baseline, except the Batch Normalization is replaced with Group Normalization(GN)~\cite{wu2018group} and Weight Standardization is applied (WS)~\cite{qiao2019weight}.
These modifications were shown to improve transfer and pre-training performance.
We study a range of sizes: the smallest is R50x1 (the same size as the baseline), and the largest is R101x3---a 101-layer ResNet, with three times wider hidden dimensions.\footnote{We do not use the largest BiT model, R152x4, since applying this massive model to high resolution images was infeasible on our available hardware for this study.}
The R50x1 has 24M parameters, and the R101x3 has 380M parameters.
These experiments are at a much larger scale then previous studies analyzing transfer for medical imaging~\cite{raghu2019transfusion, neyshabur2020being}, our smallest model (R50x1) being the largest used in those studies.

BiT models are pre-trained on three datasets of increasing size:
ILSVRC-2012 ``ImageNet'': 1.3M images, each labelled with one of 1k classes~\cite{russakovsky2015imagenet}.
ImageNet-21\textit{k}: the full ImageNet dataset containing 14M images, each labelled with one or more from 21k classes.
JFT-300M: 300M images, each with one or more noisy labels from 18k classes with an average of 1.29 labels per image.

\subsection{Imaging Tasks}
\label{sec:tasks}
We study three popular tasks: interpretation of chest X-Rays (CXR), breast cancer detection from mammograms (mammography) and diagnosing skin conditions from images taken via handheld digital devices (dermatology). For each task, we also use an alternative evaluation dataset facing some distribution shift but with an identical label space.\footnote{
All medical datasets used in this work were de-identified prior to the study. 
Note also: our data splits differ to~\cite{mckinney2020international,liu2017detecting}, so results in mammography/dermatology may differ slightly to these prior works.}

These tasks share some characteristics common to medical imaging;
the datasets are imbalanced, and pathologies are identified using features localized to small patches.
However, they also contain various unique properties.
For example, in their label structures: breast cancer detection is a binary classification problem, chest X-Ray interpretation is a multi-label problem (multiple independent binary outcomes), and skin condition diagnosis is a multi-class problem. 
Further, while the dermatology images are visually quite similar to natural images, X-Rays and mammograms are captured using specialist equipment so have very different low-level statistics.
Moreover, all these tasks operate with images at different image resolutions.
The diversity of these tasks help us ascertain the generality of the methods studied. Further details on each dataset is included in Appendix \ref{app:datasets}, but here we briefly describe them alongside the distribution-shift alternatives.

\paragraph{Mammography}
For the mammography task (Mammo), we experiment using the de-identified datasets used by McKinney \textit{et al.}~\cite{mckinney2020international}. Coming from regular screenings in the UK/US, it consists of high resolution (2048 x 2048) X-Ray images, with 4 views available per patient alongside biopsy-confirmed outcomes. These outcomes are temporally binned and projected in order to yield binary ground truths at different time horizons (i.e. if a patient had a positive biopsy at $n$ months, the ground truth is positive for all time horizons $> n$). We focus on the 39 month outcome for model evaluation.

The main dataset used for training/evaluation is the UK dataset. The US dataset is used to assess distribution shift; as well as different distributions of age/ethnicity/cancer rate of the patients, the US data has different screening practices. Previous work on these datasets\cite{mckinney2020international} demonstrated reduced performance of both radiologists and models on US data.

\paragraph{Chest X-Ray (CXR) Interpretation}
We use the CheXpert dataset~\cite{irvin2019chexpert}, a publicly available dataset of anonymized chest X-Ray images. We predict independently the five pathologies considered by Irvin \& Rajkurpar\textit{et al.}\cite{irvin2019chexpert}, treating it as a multilabel binary classification problem. We report mean AUC across these conditions. As the validation set is small, we randomly re-split the training set into a a training, validation, and test sets containing 167\,429, 22\,240 and 33\,745 datapoints, respectively.
\noindent
For distribution shift assessment, similar to previous works~\cite{rajpurkar2020chexpedition}, we use the NIH Chest-Xray8 dataset~\cite{Wang_2017}, which was collected at a different hospital/location in the US.

\paragraph{Dermatology}
For the dermatology task (Derm), the dataset (`Dataset A') consists of multiple de-identified case images taken by medical assistants in a teledermatology setting, with large variation in background, scale, perspective and lighting. 26 of the most common conditions, and an `other' class, are the primary prediction endpoint, yielding a multi-class classification problem. We use a development set of 12980 cases and a test set of 2456 cases, with up to 6 images being used for each case during training. We measure top-1 accuracy against the top label in the ground truth, collected from aggregating the assessment of multiple US-board certified dermatologists.

Distribution shifted data (`Dataset B') comes from skin cancer clinics in Australia and New Zealand. As well as a geographic shift, the dataset focuses on different conditions (cancerous conditions like Melanoma vs. non-cancerous conditions like Eczema), has ground truths from clinical biopsies, and contains different artifacts (e.g. ink markings) that mark excisions, making it a challenging setup to evaluate generalization.


\section{Experiments}


We study on the tasks outlined above whether large-scale pre-training helps model medical images. 
We study the following properties with respect to the two main features of transfer learning: pre-training dataset and model size.

\begin{itemize}[itemsep=1pt,topsep=1pt,leftmargin=12pt]
\item \textbf{In-domain Performance} To what extent does transfer improve the ``standard''  hold-out performance of the resulting model on the evaluation sets on the targets tasks? Does image resolution, which is typically large for medical images, impact transfer performance?
\item \textbf{Generalization under Distribution Shift} Do resultant models better generalize to new population distributions, such as patients from new hospitals or countries?
\item \textbf{Data Efficiency} Can we reduce the amount of (expensive) supervised medical imaging data needed to train these models to high levels of performance?
\item \textbf{Subgroup Fairness} - How do large scale pre-trained models impact fairness considerations? Could unknown bias in the pre-training data result in disproportionate benefits for some subgroups at the expense of others?
\item \textbf{Uncertainty Estimation} Safety-critical settings require that models \textit{know when they don't know}. How does pre-training at scale impact model calibration?
\end{itemize}

\section{Results}
\paragraph{Hyperparameter Tuning and Error Bars}
We run hyperparameter sweeps for each model on each task (details in Appendix~\ref{app:training_details}). We retrain multiple replicas of the best model according to the validation set performance, and then evaluate on the test set. These replicates are used to analyze the performance on task-specific endpoints, and for subgroup fairness/calibration analysis. We train five replicas for Mammography task, and ten for the Dermatology and CheXpert tasks. Error bars shown are 95\% bootstrapped confidence intervals.\\

We now discuss the effects of large scale natural pre-training on medical imaging tasks.

\subsection{Superior In-domain Performance}
\begin{figure}[t]
     \centering
     \includegraphics[width=0.9\textwidth]{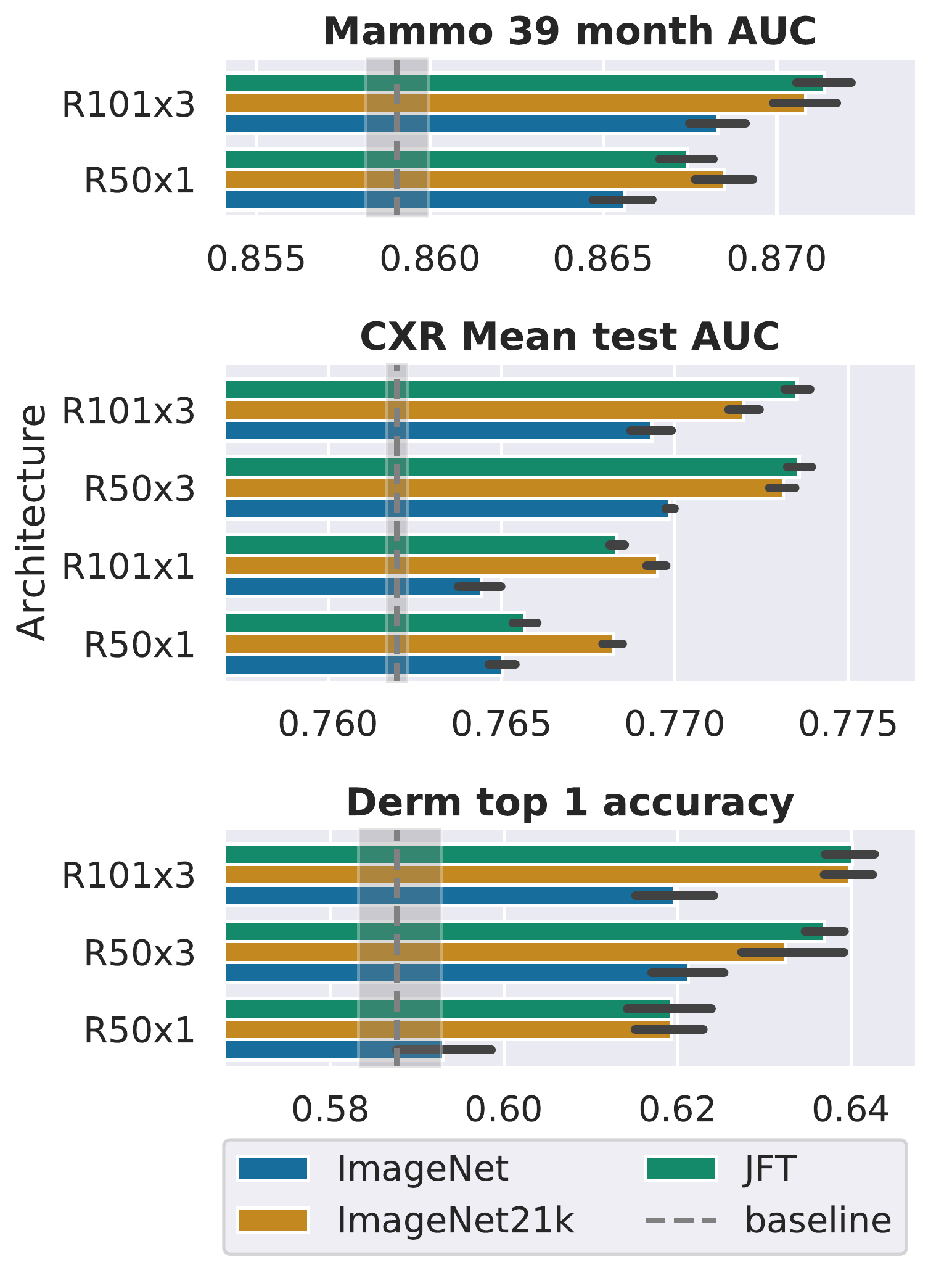}
    \caption{Transfer learning performances on the held out test set for the Mammography, CheXpert, and Dermatology tasks. On all the tasks, BiT models outperform a R50x1 baseline; generally there are benefits to scaling pre-training as well as the architecture size.}
    \label{fig:performance_gains}
\end{figure}

We first examine whether we see transfer performance improvements on the standard held-out test sets. 
Figure~\ref{fig:performance_gains} shows the results.
We observe that when pre-trained on ImageNet, the BiT R50x1 model is either on par (Dermatology) or better (Mammography, CheXpert) than the baseline R50x1 with Batch Normalization layers. 
Note, that the upstream performance on ImageNet is very similar to the baseline BiT-R50x1 model pre-trained on ImageNet alone.
This suggests that Group Normalization and Weight Standardization improve over Batch Normalization for transfer to the medical imaging setting as well.

On all the three tasks, we find both the ImageNet-21\textit{k} and JFT pre-trained BiT models improve over the ImageNet pre-trained models at all architecture sizes. 
The transfer performance on all the three tasks improves as we scale up the architecture size along with the pre-training dataset. The best performing model on the three tasks is JFT pre-trained R101x3 achieving an improvement of +1.2\% AUC, +1.2\% mean AUC and +5.3\% top-1 accuracy on the Mammography, CheXpert, and Dermatology held out test sets, respectively.
These results suggest that, at sufficient scale, both in terms of pre-training data and architecture, transfer from the natural domain can be highly effective for medical images.

Interestingly, the ImageNet-21\textit{k} pre-trained BiT models perform competitively with the JFT pre-trained ones,
whereas on natural datasets, the JFT model was reported as better~\cite{kolesnikov2019big}.
This suggests that while large improvements over ImageNet are possible, further scaling of natural datasets may have limited returns for medical applications.

\begin{figure}[t]
     \centering
     \includegraphics[width=0.96\textwidth]{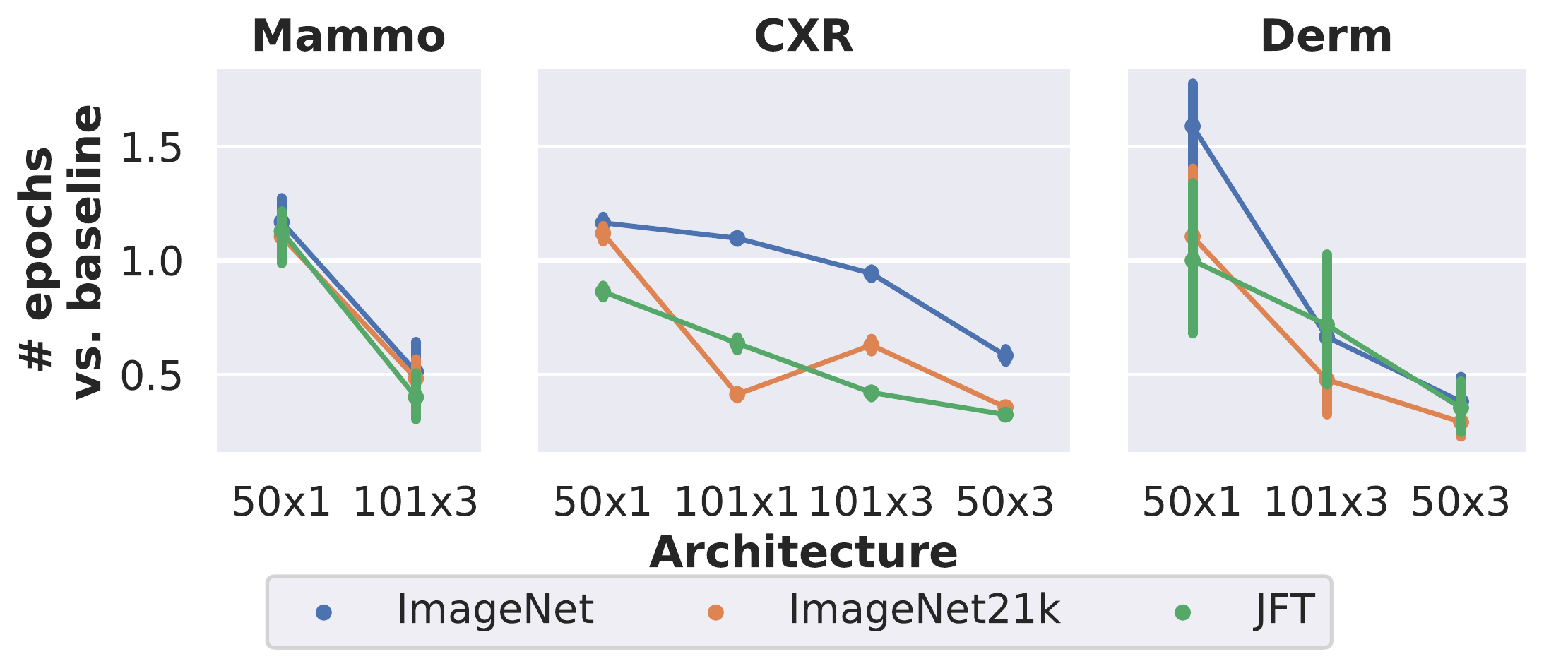}
    \caption{Number of epochs, relative to baseline, required to converge. Larger pre-trained models reduce the number of steps required to converge.}
    \label{fig:convergence}
\end{figure}

\paragraph{Faster Convergence}
Though larger scale models incur a higher compute cost, we find that this is partially offset by faster convergence on downstream tasks.
Figure~\ref{fig:convergence} clearly shows that the larger models reach highest performance faster than their smaller counterparts. These results are also in line with observations in~\cite{raghu2019transfusion, neyshabur2020being}.

\paragraph{Better Use of High-Resolution Images}
Medical images are often larger than natural images; Bit models are pre-trained only at $224\times 224$, yet medical images are sometimes at mega-pixel scale or larger.
We fine-tune models at a range of resolutions and test whether pre-training continues to yield performance gains at higher resolutions. For all modalities, we train and evaluate models at three resolutions using the best parameters from the original hyperparameter sweep.
Perhaps surprisingly, relative performance improvements compared to the baseline models tended to increase with resolutions for both Dermatology and Mammography.
The large resolution mismatch does not seem to impede performance improvements due to transfer to these tasks.
There was no clear trend for CheXpert, although absolute performance on CheXpert did not change much with increased resolution, indicating that little task-relevant information is gained when training at higher resolution. This may be due to the labels in CheXpert relying on large image features, e.g. cardiomegaly or consolidation both being relatively large features of an X-Ray.
In contrast, cancers typically form a very small part of a mammogram, and dermatological features are often highly localized.

\begin{figure}[t]
    \centering
    \includegraphics[width=0.8\textwidth]{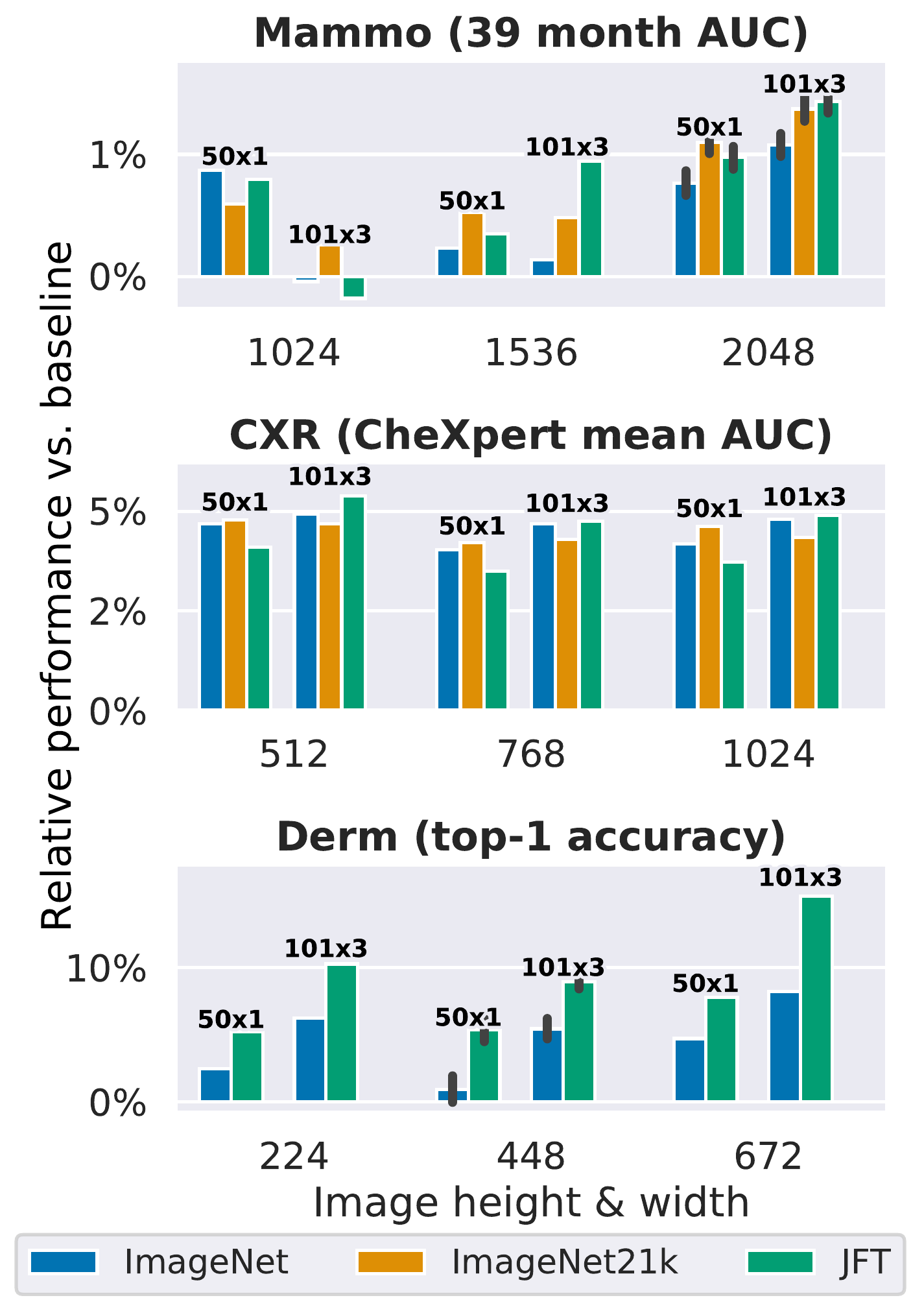}
    \caption{
    Relative performance compared to the baseline model at various resolutions.
    For Mammography and Dermatology, larger models tend to yield more improvements at larger resolution---note the baseline also improves with resolution (not visible in the figure).
    For CheXpert, within the range tried, performance was independent of resolution.\vspace{-1mm}} 
    \label{fig:resolution-result}
\end{figure}

\paragraph{Summary}
We show with sufficient data and architecture scale, transfer can indeed be effective on medical imaging tasks.
Fewer training steps are needed for larger models, despite the domain shift, and perhaps surprisingly, they work at high image resolutions seen in medical imaging.

\subsection{Robust Generalization under Distribution Shift}
Handling domain shift, arising from deployment in new hospitals or geographies, changing disease prevalence and so on is key to successful deployment in clinical settings.
Since distribution shifts may be unknown in advance of deployment, we do not fine-tune the models on the new distribution or make any modifications to training on the original datasets to account for it.
For each modality, we evaluate trained models on alternative domains, but with the same class label space.\footnote{Note: the original training data for Mammography included US data. We retrain 3 models on only UK data with the best hyperparameters, and evaluate on US data.}
The specific domain shifts present in each task are described in Section~\ref{sec:tasks}.

\begin{figure}[t]
    \centering
     \includegraphics[width=0.8\textwidth]{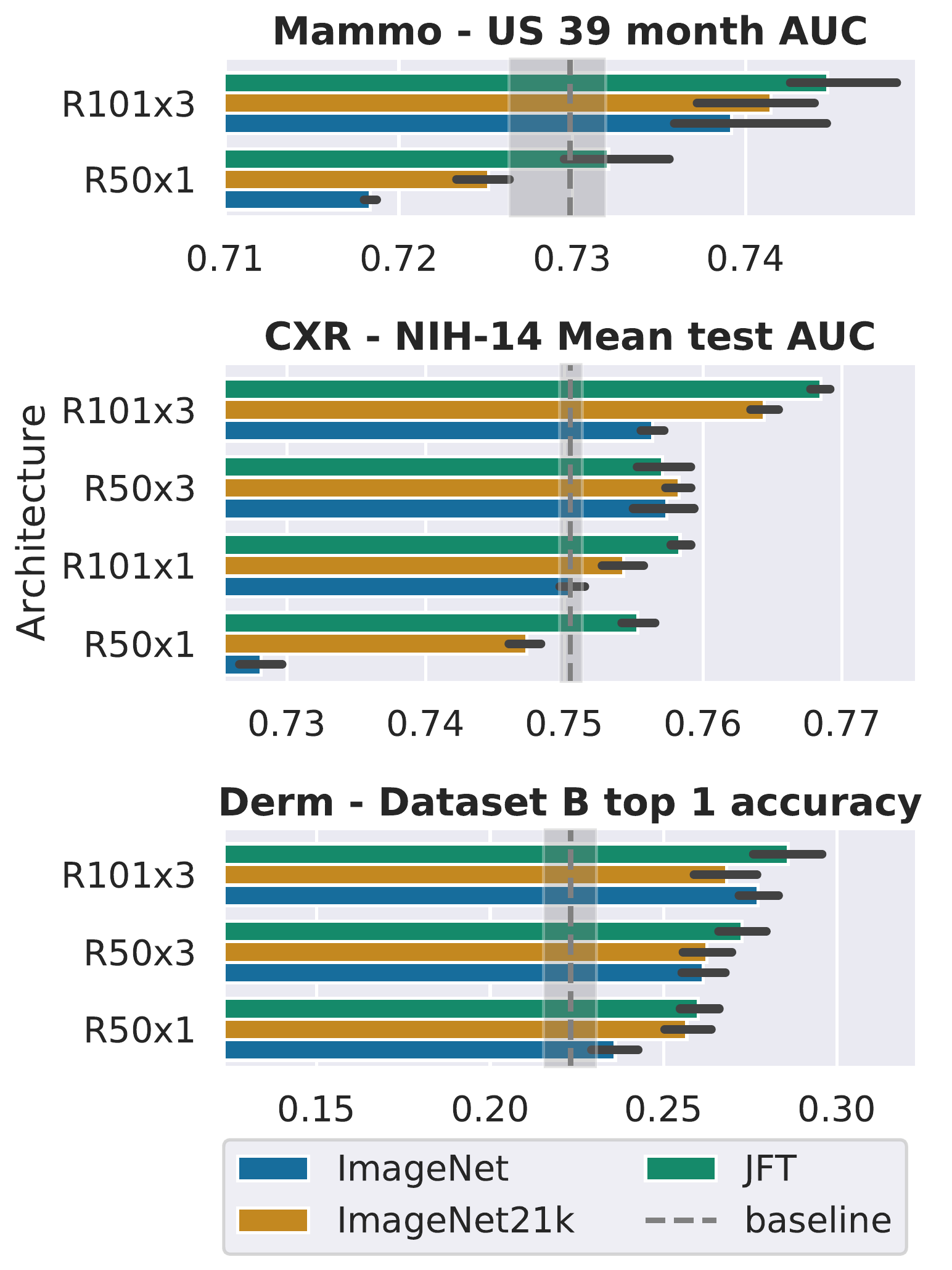}
    \caption{Generalization of models to other datasets in the same domain. Increasing pre-training scale and architecture scale generally improves robustness to distribution shift on all the tasks considered.}
    \label{fig:zero_shot}
\end{figure}

The results are shown in Figure~\ref{fig:zero_shot}, with full numerical results in Table~\ref{tab:zero_shot}, Appendix~\ref{app:tables}.
Across the board, increasing architecture and pre-training data scale improves robustness to domain shift. Compared to the baseline, the JFT pre-trained R101x3 achieves absolute improvements of +1.5\% AUC, +1.9\% mean AUC, +6.3\% top-1 accuracy on the alternate datasets for Mammography, CheXpert, and Dermatology tasks. Comparing to the ImageNet pre-trained R101x3 BiT model, 0.6\%, 1.3\% and 0.9\% of this improvement was due to the scale of the pre-training data.

The smaller BiT models---particularly the ImageNet pre-trained BiT-R50x1---do not generalise as well to the new datasets. These results follow \cite{djolonga2020robustness}, albeit on natural tasks and further consolidate the importance of scaling up both architecture size and pre-training data for training robust representations for transfer to medical imaging tasks.

The differences in model performance observed here are much more pronounced than the differences in transfer performance on the original held out test set; for example, the best and worst CheXpert models have a performance difference of approximately 1\% on the CheXpert test set, but over 4\% on the NIH dataset. This is also an indication of the underspecification phenomenon~\cite{damour2020underspecification} where models with equivalent test performance may still exhibit vastly different behaviors on stress tests or in deployment settings.

\paragraph{Summary}
Scaled transfer learning not only improves in domain performance, but also guards against performance degradation due to distribution shift. This finding is key for deployment in real world clinical settings.

\subsection{Improved Data Efficiency}
Large scale pre-training has demonstrated impressive results on low-data tasks in vision and NLP~\cite{kolesnikov2019big,brown2020language}. 
In the natural domain, BiT achieves competitive performance in few-shot classification using simple fine-tuning. 
We aim to see if scale similarly improves data efficiency in the medical setting where labelled data is often a bottleneck.
For this, we reduce the amount of task-specific data for fine-tuning. Classes are separately sub-sampled to keep class balance consistent at different dataset sizes.
For computational reasons, we use the best model setting from the hyperparameter sweep from the full-data experiments for all subsets.

\begin{figure}[t]
     \centering
     \includegraphics[width=0.9\textwidth]{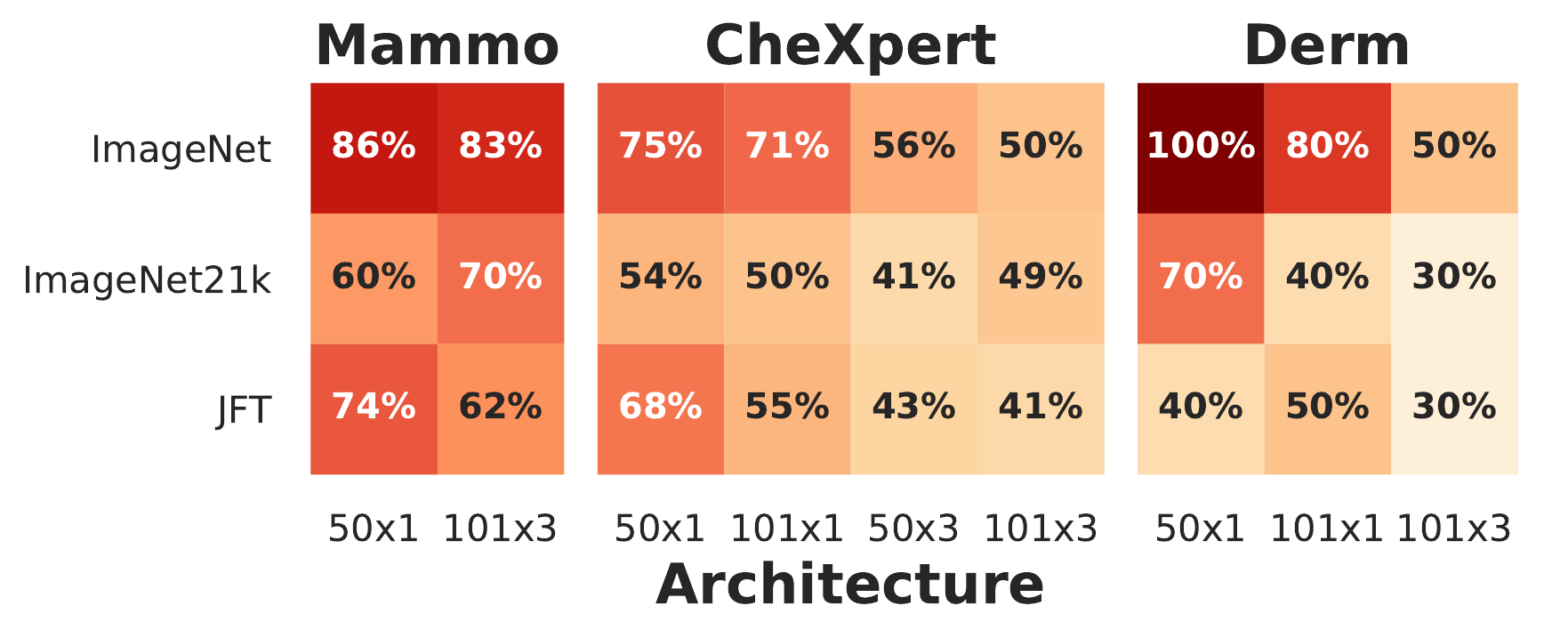}
    \caption{Data required to reach baseline performance. Improvements from increasing architecture and pre-training scale are conserved at lower data regimes.}
    \label{fig:data_for_baseline_performance}
\end{figure}

Figure~\ref{fig:data_for_baseline_performance} shows the fraction of the training data required to match baseline performance. 
We find that, larger models seem to require less data even when transferring to medical images.
The R101x3 achieves the baseline performance on the Mammography, CheXpert, and Dermatology tasks with 62\%, 41\% and 30\% of the data, respectively. 

Analysing the \textit{rate} at which performance decays with respect to the amount of data reveals some nuances.
Gains from scaling up architecture and pre-training are conserved at all data fractions. However, when reducing data, the \textit{relative} drop in performance is roughly consistent across all architectures/initialization considered, including the baseline. We therefore believe that improvements in data efficiency in this sense necessitate more specialised medical data for pre-training. The full curves of performance vs. amount of data are shown in Appendix~\ref{app:data_efficiency}


\paragraph{Summary} We reduce one of the most significant burdens in medical imaging research---acquiring data---by reaching baseline performance with 40\% to 70\% less  data, depending on the task.

\subsection{Equivalent Subgroup Fairness}
It is possible that harmful biases from the pre-training stage could be carried forward to downstream tasks; even unsupervised pre-training has been shown to pick up (some harmful) human biases~\cite{steed2020unsupervised}. We therefore analyse performance of models on subgroups, using breast density, age and skin tone as the protected characteristic for the Mammography, CheXpert, and Dermatology tasks, respectively. Further analyses are shown in Appendix~\ref{app:fairness}, but in Figure~\ref{fig:fairness_deviations} we show the gap between the worst performing subgroup and the mean performance. Note that this still penalises models which yield improvements in all subgroups if the performances then become more uneven.

While we notice no systematic trend between pre-training/architecture scale and performance across subgroups, it is interesting to note that the choice of initialisation and architecture does impact different subgroups differently. Thus, when practitioners make decisions they should be aware it can have unintended side effects if they simply optimise for aggregate metrics. We leave for future work to see whether, when explicitly optimised for, large scale transfer can yield better trade-offs than other approaches. 

\paragraph{Summary} A key concern when using transfer learning---exacerbating biases due to biased pre-training data---does not materialise in the dimensions we consider. 

\begin{figure}[t]
     \centering
     \includegraphics[width=1.0\textwidth]{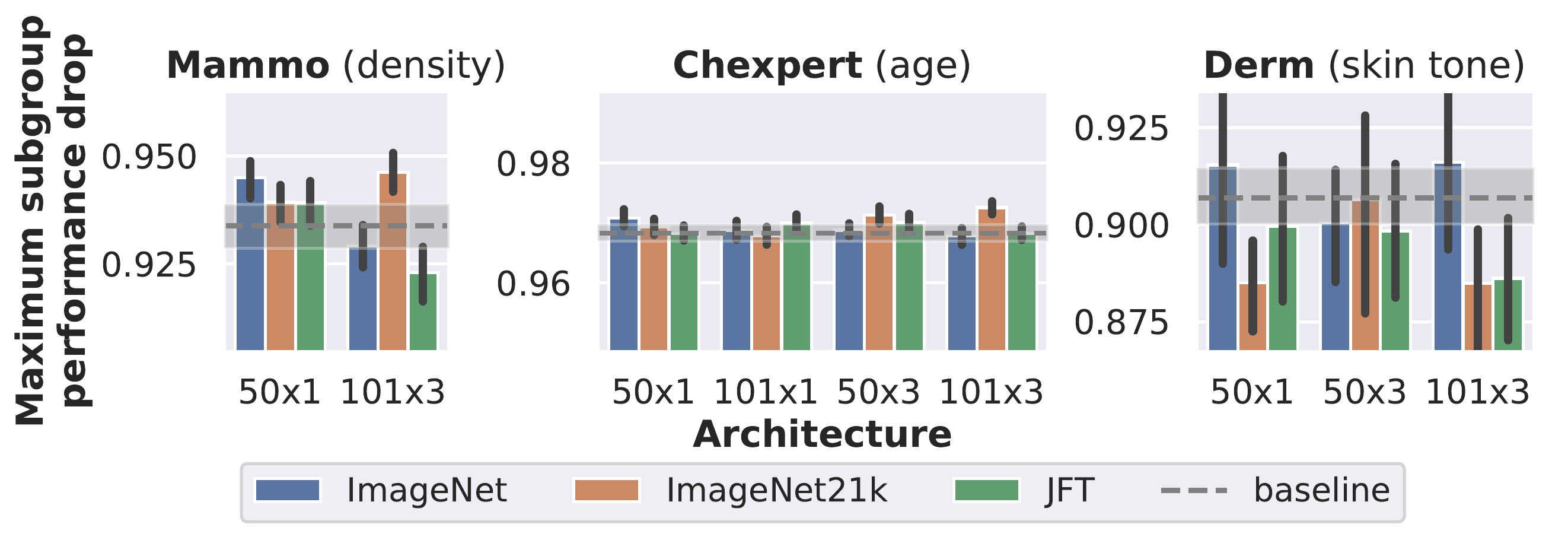}
    \caption{Largest relative deviation from mean performance between subgroups. Choice of pre-training dataset/architecture can clearly affect different subgroups differently, but we do not observe a systematic relationship.}
    \label{fig:fairness_deviations}

\end{figure}
\subsection{No Sacrifice in Uncertainty Estimation}
\begin{figure}[t]
     \centering
     \includegraphics[width=1.0\textwidth, height=0.35\textwidth]{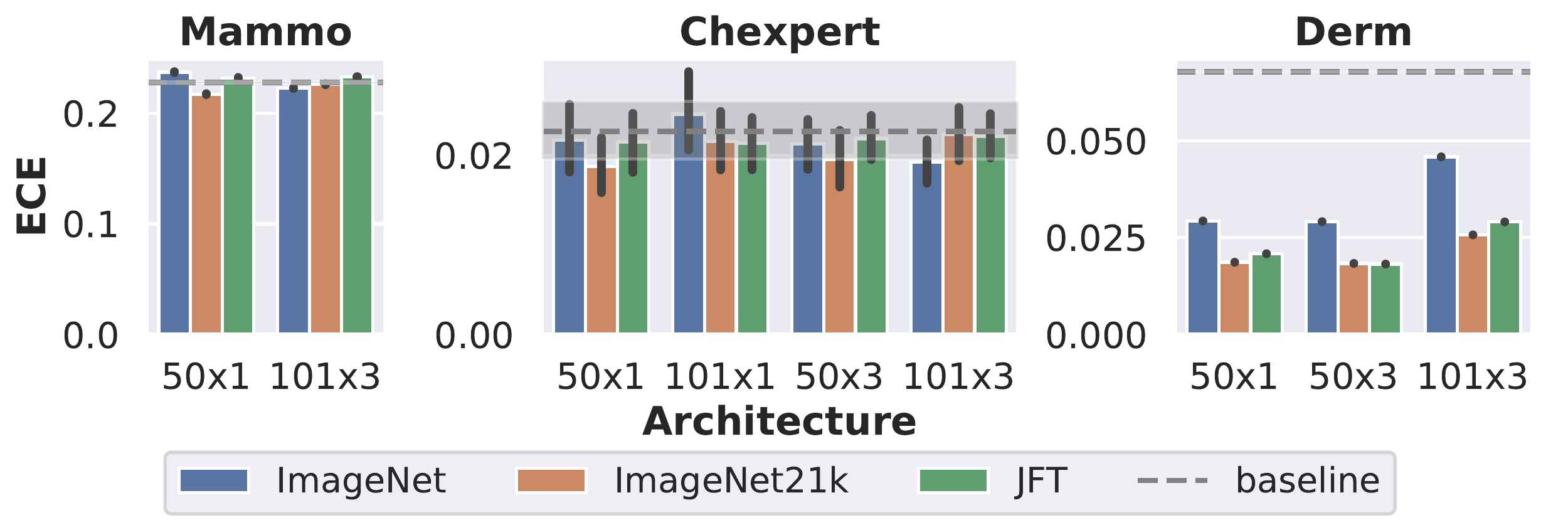}
    \caption{Expected calibration errors for different models. Increased pre-training dataset scale improves calibration for Dermatology, but otherwise there is no systematic trend.}
    \label{fig:eces}
\end{figure}

Previous studies have shown that larger deep neural networks can be more poorly calibrated~\cite{guo2017calibration}.
However, research on calibration tends to focus on natural tasks, such as CIFAR and ImageNet.
Therefore, it is unclear what effect large scale transfer would have on medical images.
Figure~\ref{fig:eces} shows Expected Calibration Errors (ECE) for our trained models.
For CheXpert and Mammography we found that the model scale and pre-training had little effect on calibration in this regime. For Dermatology---where all models are very well calibrated anyway---the BiT models are better calibrated than the baseline, and JFT/ImageNet-21\textit{k} models are always better than ImageNet. There is a minor increase in calibration due to scaling the architecture, but this is offset by the improvement from scaling pre-training.
The difference in calibration between the tasks is much greater than the difference between the calibration of the models.
Other modelling choices, such as the use of focal loss for the Mammography model and the smooth probabilistic labels (obtained from averaged dermatologist ratings) in dermatology, seem to influence the calibration more strongly.

\paragraph{Summary} There appears to be either mild benefit or little influence of large scale transfer on calibration for medical images.

\subsection{Understanding the models}
\begin{figure}[t]
\floatbox[{\capbeside\thisfloatsetup{capbesideposition={right,top},capbesidewidth=0.4\textwidth}}]{figure}[\FBwidth]
{\hspace{-0.5cm}\caption{Average movement in different ResNet blocks of Mammography models. For models pre-trained with larger datasets, earlier kernels are modified more, but deeper ones are modified less.}\label{fig:mammo_blocks}}
{\includegraphics[width=0.55\textwidth]{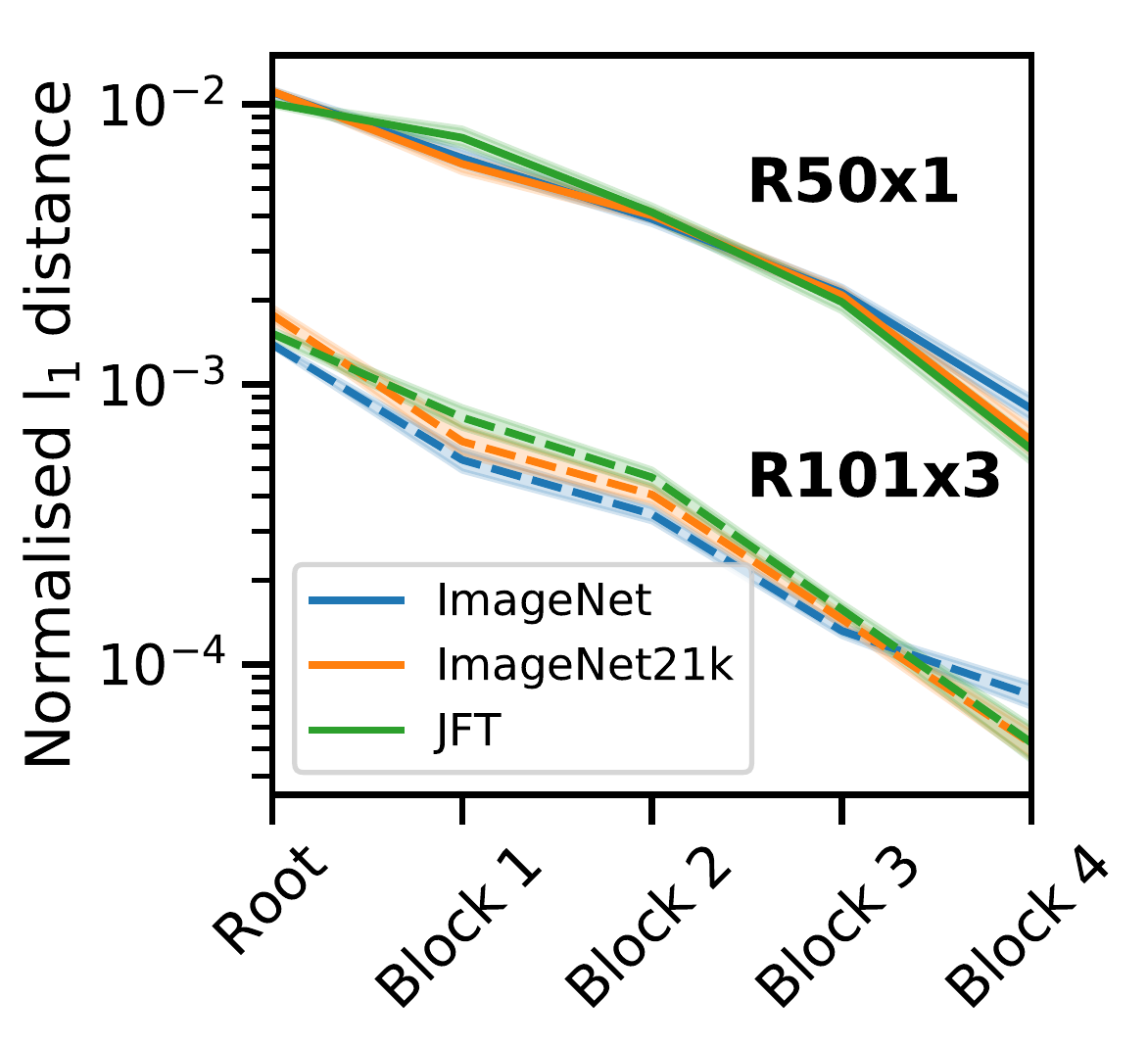}}

\end{figure}

\begin{figure}[t]
     \centering
     \includegraphics[width=0.9\textwidth]{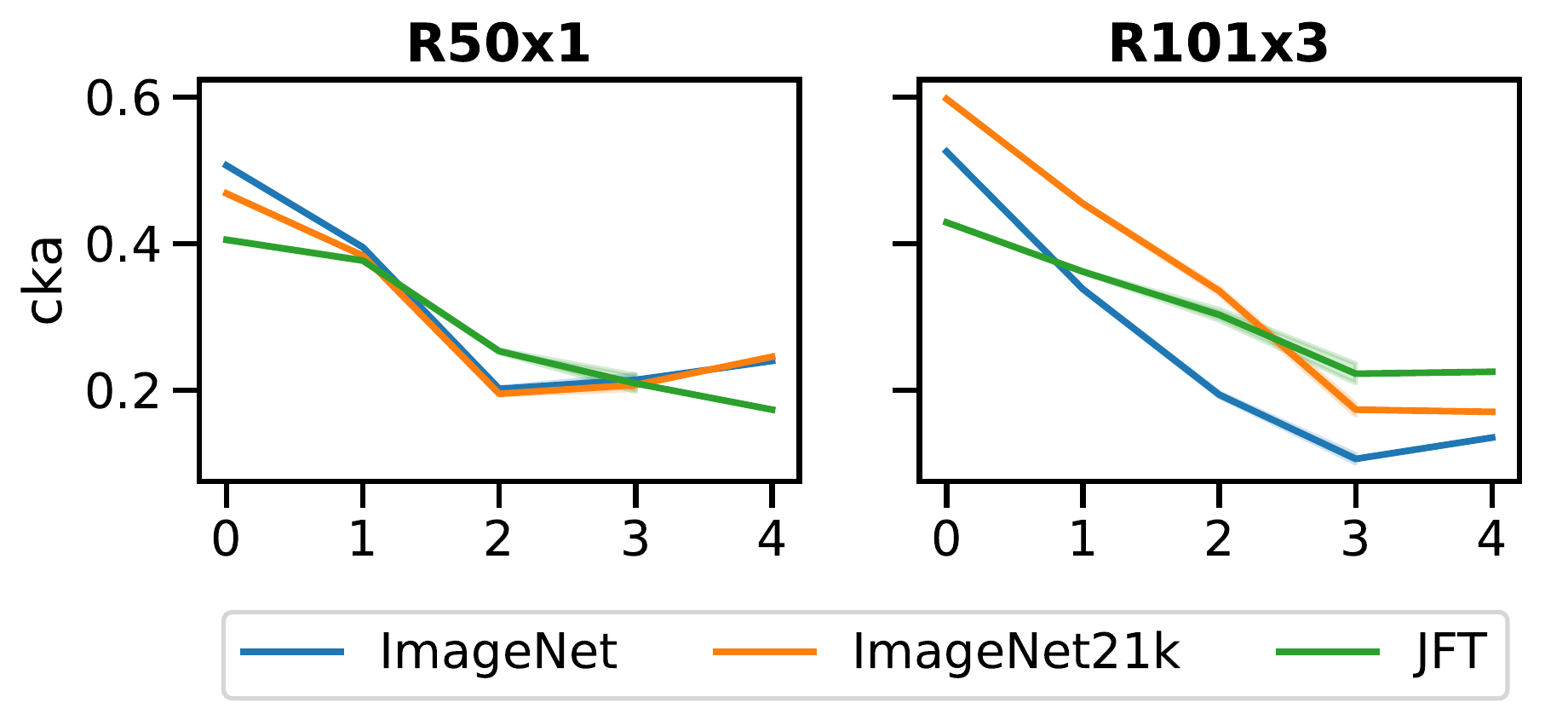}
    \caption{Scale allows models to better use higher level features. Linear CKA shown between initialisation and final dermatology models.}
    \label{fig:derm_cka}
\end{figure}

To discern the source of shown benefits, we analyse the weights and features of the trained models. We use linear centred kernel alignment (CKA)~\cite{kornblith2019similarity} for the latter.
Previous works~\cite{raghu2019transfusion,neyshabur2020being} have shown that transferred models often struggle to re-use higher level features.
Figure \ref{fig:derm_cka} shows the CKA between initialisation and the trained dermatology model at different depths. For the larger architectures, the deeper features are more similar to initalization for JFT an ImageNet-21\textit{k}; this may indicate that the combination of architecture scale \textit{and} pre-training scale allows better use of high-level features. Figure~\ref{fig:mammo_blocks} shows the average movement in parameter space of each ResNet block. Models pre-trained with larger datasets exhibit larger weight changes in the earlier layers, but smaller ones in the later layers, further indicating possible re-use of higher level features.
Further insights are available in Appendix~\ref{app:weight_feature_analysis}, where for example we show that larger architectures move less in parameter space.
This correlates with the observed convergence speedup here and in other works~\cite{raghu2019transfusion}.

Finally, we analyse qualitative examples for the Mammography model as a sanity check to ensure that the scaling allows modesls to pick up meaningful features and not just better fit to spurious correlations. These are shown in Appendix~\ref{app:examples} and show that the improved accuracy corresponds to identifiable abnormalities in the images.

\section{Conclusion}
We have studied the benefits of large-scale supervised pre-training for medical imaging applications. 
We show that using larger architectures and larger pre-training datasets can yield significant improvements across a diverse range of medical domains, in spite of significant difference from the pre-training data. 
As well as improvements to performance and data efficiency, we show particularly strong improvements to generalization under distribution shift, and show that these improvements do not come at the cost of subgroup fairness or model calibration. 
Note, our findings are orthogonal to alternative approaches to pre-training for medical imaging, such as use of in-domain labeled data or self-supervised learning. 

The ImageNet-21\textit{k} models used in this study are open-sourced and easily accessible\footnote{
\url{https://tfhub.dev/google/collections/bit/1}
\url{https://github.com/google-research/big\_transfer}},
and to a large extent perform similarly to the JFT pre-trained models.
We hope, therefore, that practitioners find it useful to at least migrate from standard ImageNet checkpoints to capture the benefits we have demonostrated here.

\ifcvprfinal
\section{Acknowledgments}
We would like to thank Xiaohua Zhai for detailed and valuable feedback on the manuscript. We would also like to thank Varun Godbole, Sylvain Gelly, Jay Hartford, Shravya Shetty, Dale Webster, Greg Corrado and our collaborators at Northwestern Medicine, DermPath AI, Cancer Research UK; the OPTIMAM project team and staff at the Royal Surrey County Hospital who developed the UK mammography imaging database and all members of the Etemadi Research Group for their support of this work.  
\fi
\newpage

\small
\bibliographystyle{ieee_fullname}
\bibliography{main}

\appendix
\clearpage
\section{Datasets}
\label{app:datasets}
\subsection{Mammography}
\textbf{Main dataset}: We use the UK dataset of \cite{mckinney2020international}; it was collected as part of routine screening programs for women aged 50 and 70 across three hospitals. In total, the training data contained 53\,152 negative cases and 7\,571 positive cases. We used a dataset containing 12\,309 negative cases and 246 positive cases for hyperparameter selection (the validation set), and report final performance on a dataset of 48\,933 negative and 1\,003 positive cases (the test set). Note that these counts are different from those reported in \cite{mckinney2020international}, because we re-split the training set to create an additional validation set for tuning. Cases were divided into the datasets at the patient level.

\textbf{Distribution-shift dataset}: US Dataset
 To test generalisation under distribution shift in the mammography, we use a dataset collected in a different country (USA), and with slightly different screening criteria. The dataset is the same as that discussed in \cite{mckinney2020international}. Data was collected between 2001 and 2018 from 3,097 women aged 40 and upwards assessed at Northwestern Memorial Hospital (US). We included images from all 1,511 women who were biopsied during this time period and a random subset of women who never underwent biopsy (Methods). Among the women who received a biopsy, 686 were diagnosed with cancer within 27 months of imaging.

\subsection{Chest X-Ray}
The CheXpert dataset \cite{irvin2019chexpert} is a publicly available dataset of anonymized chest x-ray images. The images, alongside radiology reports, were collected retrospectively. Labels were extracted from the reports automatically, and consist of 14 radiological findings, categorised into `not mentioned', `negative', `positive', and `uncertain'. As is commonplace, we map `uncertain' to `positive', and `not mentioned' to `negative' and treat it as multilabel binary classification. The training dataset consists of 224\,316 chest radiographs of 65\,240 patients. Due to the small size of validation set (234 manually annotated x-rays), we split the training set into a a training, validation, and test sets containing 167\,429, 22\,240 and 33\,745 datapoints, respectively; we found comparing models on the validation set unreliable due to the small size.

\textbf{Distribution-shift dataset}: NIH Chest X-Ray
The alternative dataset used for CXR was the ChestX-ray8 dataset \cite{Wang_2017}. The dataset contains more than 100k deidentified chest x-ray images collected at a different hospital. Labels for common radiological findings were constructed using text-mining of the radiological reports. The dataset contains labels for Atelectasis, Cardiomegaly, Effusion, Infiltration, Mass, Nodule, Pneumonia, Pneumothorax.

\subsection{Dermatology}
\textbf{Main dataset}: ``Dataset A'' comes from a retrospective dataset with anonymized data. The reference standard for the cases was collected from one or more dermatologists, who provided differential diagnoses for each case. The images were taken by medical assistants, with large variation in background, scale, perspective and lighting. Cases with multiple skin conditions, or that were not diagnosable were excluded, resulting in a dataset of 15\,436 cases (76\,478 images). 
The dataset consists of 288 skin conditions, of which we identified 26 as the most common skin conditions that served as the primary endpoint for our prediction. The remaining conditions were grouped under a separate ‘Other’ class.
We split the data into a development and test set, using a temporal split, such that cases in the development set came before cases in the test set. We ensured that no patient appeared in both sets. This results in a development set of 12\,980 cases and a test set of 2\,456 cases, with up to 6 images being used for each case during training. Training and hyperparameter selection was done solely using cases in the development set, which we further split into train and tune splits, and we report final performance numbers on the test set.

\textbf{Distribution-shift dataset}: Australia/New Zealand Dataset

Cases in the out-of-distribution dataset for Dermatology were collected from a chain of skin cancer clinics in Australia and New Zealand. We refer to this as 'Dataset B' throughout. This dataset is primarily focused on skin cancer, covering a wide variety of cancerous conditions such as Melanoma, Basal Cell Carcinoma, and Actinic Keratosis. This dataset differs from the Dataset A used in the other transfer experiments on several important dimensions. Firstly, the ground truth for the examples come from biopsies, instead of the aggregated assessments of dermatologists. Secondly, there is a marked distribution shift in conditions, with cancerous conditions being the most prevalent. In contrast, the top conditions in the Dataset A are Eczema and Psoriasis, both of which are non cancerous. Thirdly, because of the nature of the cases, some of the zero shot images may contain artifacts such as ink markings that mark the site of excisions. These add up to form a difficult dataset to do zero-shot transfer on.

\section{Training details}
\subsection{Modelling details}
\label{app:modelling}
\paragraph{Mammography}
We use the global case-wise prediction model of \cite{mckinney2020international}. Features extracted from each view are concatenated to yield a case-wise representation. For each time horizon, this representation is passed through a 512-dimension ReLU activated hidden layer with a dropout rate of 0.25. A sigmoid-activated 1-dimensional layer yields a binary probability of cancer developing in that time horizon. We predict five time horizons during training (3, 12, 15, 27 and 39 month).

As in the original setup, two measures are included to handle data imbalance: First, positive cases are sampled more frequently to give roughly equal presence during training. Second, a focal loss \cite{lin2017focal} with $\gamma = 1.5$ is used. Note that parameters for dropout rate, SGD momentum and the focal loss were heavily tuned for the baseline model; in our efforts, we used a simple fixed hyperparameter search space over the learning schedule.

\paragraph{Chest X-Ray}
Features were extracted independently from each X-ray image, and dropout of rate 0.2 was applied. The resultant representation was passed through a single dense layer followed by a sigmoid activation to yield 5 independent binary predictions for each condition. We do not combine in any way multiple views/scans for the same patient.

\paragraph{Dermatology}
Features are extracted independently for each dermatology image within a case, and then averaged to give a final case-wise representation. This is then passed through a fully-connected layer. Dropout of rate 0.2 is applied, followed by softmax to yield probabilities over the output label space. This effectively produces a ranked list of conditions which form the differential diagnosis of the model. The dataset and model setup follows \cite{liu2020deep}.

\subsection{Training setup and hyperparameter sweep}
\label{app:training_details}
For all models, we used an exponential decay step schedule with warmup; the learning rate was linearly warmed-up from zero to some \texttt{initial\_learning\_rate}. At every subsequent \texttt{decay\_steps}, the learning rate was decayed by a factor of \texttt{decay\_factor}.
\subsubsection{Mammography}
Mammography models are trained on Google Cloud TPUv3s. The baseline, and BiT-R50x1 models were trained with a batch size of 16, and BiT R101x3 models were trained with a batch size of 8. Model hyperparameters were predominately kept identical to \cite{mckinney2020international}. 
\paragraph{Hyperparameter sweep for baseline \& R50x1 models}
\begin{itemize}
    \item \texttt{schedule\_length} $\in$ \{30k, 50k\}
    \item \texttt{initial\_learning\_rate} $\in$ \{0.003, 0.001, 0.0003\}
    \item \texttt{decay\_factor} $\in$ \{0.1, 0.5\}
    \item \texttt{decay\_steps} $\in$ \{5k, 10k, 25k\}
\end{itemize}
\noindent
\paragraph{Hyperparameter sweep for R101x3 models}
\begin{itemize}
    \item \texttt{initial\_learning\_rate} $\in$ \{0.001, 0.0005, 0.00001\}
    \item \texttt{decay\_factor} $\in$ \{0.1, 0.5\}
    \item \texttt{decay\_steps} $\in$ \{10k, 25k\}
\end{itemize}

\noindent
\paragraph{Preprocessing \& data augmentation}
Images were elastically deformed using the DeepMind multidimensional image augmentation library (https://github.com/deepmind/multidim-image-augmentation).
\begin{itemize}
    \item Resized to 2048x2048 pixels, using bi-linear resampling
    \item A regularly spaced (257 pixel distance) control grid was placed on the image
    \item This grid was non-rigidly transformed by shifting each grid node by between -50 and 50 pixels in both the x and y direction, with a random angle (+-25 radians) offset
    \item The image was randomly and independently rescaled in the x and y axis by between 90 and 110\%
    \item The image was randomly flipped up/down and left/right
    \item We randomly swapped the laterality of the images, switching the RMLO/LMLO and RCC/LCC images respectively
\end{itemize}

\subsubsection{CheXpert}
These models were also trained on Google Cloud TPUv3s, with batch size 256. Learning rates were warmed up linearly for 1 epoch, then decayed by a factor of 0.1 at the 2nd and 4th epoch. Models were trained for 50 epochs in total, but typically converged long before hand.\\
Each model was tuned with the following hyperparameter sweep:
\begin{itemize}
    \item Weight decay to init $\in$ \{1E-6, 1E-5, 1E-4\}
    \item Optimizer $\in$ \{\texttt{adam}, \texttt{sgd}\}
    \item Learning rate $\in$ \{0.1, 0.03, 0.01, 0.003, 0.001\} (for \texttt{sgd}) or \{1E-3, 1E-4, 1E-5\} (for \texttt{adam})
\end{itemize}

\noindent
\textbf{Data augmentation}\\
As seems to be the case in previous works, we found the models overfit very quickly. Heavy data augmentation in this case helped prevent overfitting and improve the final performance.
At training time, the following preprocessing was applied:
\begin{itemize}
    \item Random rotation by angle $\delta \sim$ \texttt{Uniform}(-20, 20)
    \item Random crop to 224 by 224 pixels
    \item Random left-right flip with probability 50\%
    \item Linearly rescales value range from [0, 255] to [-1, 1] then:
    \begin{itemize}
        \item Random additive brightness modulation:
            Adds to all channels some $\delta \sim$ \texttt{Uniform}(-0.2, 0.2).
        \item Random multiplicative contrast modulation:
        Multiplies per-channel standard deviation by a factor $s \sim $ \texttt{Uniform}(-0.2, 0.2).
        \item Reclips values to [-1, 1] range.
    \end{itemize}
\end{itemize}

\subsubsection{Dermatology}
\textbf{Hyperparameter sweeps}\\
For all models, we used the Adam optimizer, with a sweep across learning rates. We swept across a range of 16 learning rates ranging from 1e-5 to 1e-3, logarithmically spaced.

\textbf{Preprocessing \& data augmentation}\\
For all models, we applied the following augmentations:
\begin{itemize}
    \item Random rotation by angle $\delta \sim$ \texttt{Uniform}(-30, 30)
    \item Resize to 448 by 448 pixels
    \item Random left-right flip with probability 50\%
    \item Random up-down flip with probability 50\%
    \item Random brightness deviations by a factor of \texttt{Uniform}(-0.1, 0.1)
    \item Random contrast deviations by a factor of \texttt{Uniform}(-0.2, 0.2)
    \item Random saturation deviations by a factor of \texttt{Uniform}(-0.2, 0.2)
    \item Random hue deviations by a factor of \texttt{Uniform}(-0.02, 0.02)
    \item Random blur with a 2d Gaussian kernel, with standard deviation randomly chosen from the set \{0.001, 0.01, 0.1, 1.0, 3.0, 5.0, 7.0\}
\end{itemize}
\section{Further results and analysis}
\subsection{Tabular results}
\label{app:tables}
We include here tabulated results of different experiments. Table~\ref{tab:performance_gains} shows the performances of different models on the test set of our training datasets, and Table~\ref{tab:zero_shot} shows the zero-shot generalisation under distribution shift.
\begin{table*}[ht]
\begin{tabular}{@{}lllll@{}}
\toprule
 &  & Derm & CXR & Mammo \\ 
Architecture & Pretraining & Top-1 accuracy & Mean test AUC & Test AUC \\ \midrule
Baseline & ImageNet & 58.7\rng{58.3}{59.2} & 76.2\rng{76.2}{76.2} & 85.9\rng{85.8}{86.0} \\
 & ImageNet & 59.3\rng{58.8}{59.9} & 76.5\rng{76.5}{76.5} & 86.6\rng{86.5}{86.6} \\
R50x1 & ImageNet21k & 61.9\rng{61.5}{62.3} & 76.8\rng{76.8}{76.8} & 86.8\rng{86.8}{86.9} \\
 & JFT & 61.9\rng{61.4}{62.4} & 76.6\rng{76.5}{76.6} & 86.7\rng{86.7}{86.8} \\ \midrule
 & ImageNet & - & 76.4\rng{76.4}{76.5} & - \\
R101x1 & ImageNet21k & - & 76.9\rng{76.9}{77.0} & - \\
 & JFT & - & 76.8\rng{76.8}{76.9} & - \\ \midrule
 & ImageNet & 62.1\rng{61.7}{62.5} & 77.0\rng{77.0}{77.0} & - \\
R50x3 & ImageNet21k & 63.2\rng{62.7}{63.9} & 77.3\rng{77.3}{77.3} & - \\
 & JFT & 63.7\rng{63.5}{63.9} & 77.4\rng{77.3}{77.4} & - \\ \midrule
 & ImageNet & 62.0\rng{61.5}{62.5} & 76.9\rng{76.9}{77.0} & 86.8\rng{86.8}{86.9} \\
R101x3 & ImageNet21k & 64.0\rng{63.7}{64.2} & 77.2\rng{77.2}{77.2} & 87.1\rng{87.0}{87.2} \\
 & JFT & 64.0\rng{63.7}{64.3} & 77.4\rng{77.3}{77.4} & 87.1\rng{87.1}{87.2} \\ \bottomrule
\end{tabular}
\caption{Performance improvements on held-out test set. Models were tested on data coming from the same distribution as the training data.}
\label{tab:performance_gains}
\end{table*}

\begin{table*}[ht]
\begin{tabular}{@{}lllll@{}}
\toprule
 &  & Derm & CXR & Mammo \\
Architecture & Pretraining & Dataset B Top-1 & NIH test AUC & US Test AUC \\  \midrule
Baseline & ImageNet & 22.3\rng{21.5}{23.1} & 75.0\rng{75.0}{75.1} & 73.0\rng{72.6}{73.2} \\
 & ImageNet & 23.6\rng{22.9}{24.2} & 72.8\rng{72.7}{73.0} & 71.8\rng{71.8}{71.9} \\
R50x1 & ImageNet21k & 25.7\rng{25.0}{26.3} & 74.7\rng{74.6}{74.8} & 72.5\rng{72.3}{72.6} \\
 & JFT & 26.0\rng{25.4}{26.6} & 75.5\rng{75.4}{75.7} & 73.2\rng{73.0}{73.6} \\ \midrule
 & ImageNet & - & 75.1\rng{75.0}{75.1} & - \\
R101x1 & ImageNet21k & - & 75.4\rng{75.3}{75.6} & - \\
 & JFT & - & 75.8\rng{75.8}{75.9} & - \\ \midrule
 & ImageNet & 26.1\rng{25.5}{26.8} & 75.8\rng{75.5}{75.9} & - \\
R50x3 & ImageNet21k & 26.2\rng{25.6}{27.0} & 75.8\rng{75.7}{75.9} & - \\
 & JFT & 27.2\rng{26.6}{27.9} & 75.7\rng{75.5}{75.9} & - \\ \midrule
 & ImageNet & 27.7\rng{27.2}{28.3} & 75.6\rng{75.6}{75.7} & 73.9\rng{73.6}{74.5} \\
R101x3 & ImageNet21k & 26.8\rng{25.9}{27.7} & 76.4\rng{76.3}{76.5} & 74.1\rng{73.7}{74.4} \\
 & JFT & 28.6\rng{27.6}{29.5} & 76.9\rng{76.8}{76.9} & 74.5\rng{74.3}{74.9} \\ \bottomrule
\end{tabular}
\caption{Performance under distribution shift. Larger models, pretrained on larger datasets, exhibit consistently higher performance.}
\label{tab:zero_shot}
\end{table*}

\subsection{Data efficiency}
\label{app:data_efficiency}
\begin{figure*}
     \centering
     \begin{subfigure}[b]{0.4\textwidth}
         \centering
         \includegraphics[height=1.7in]{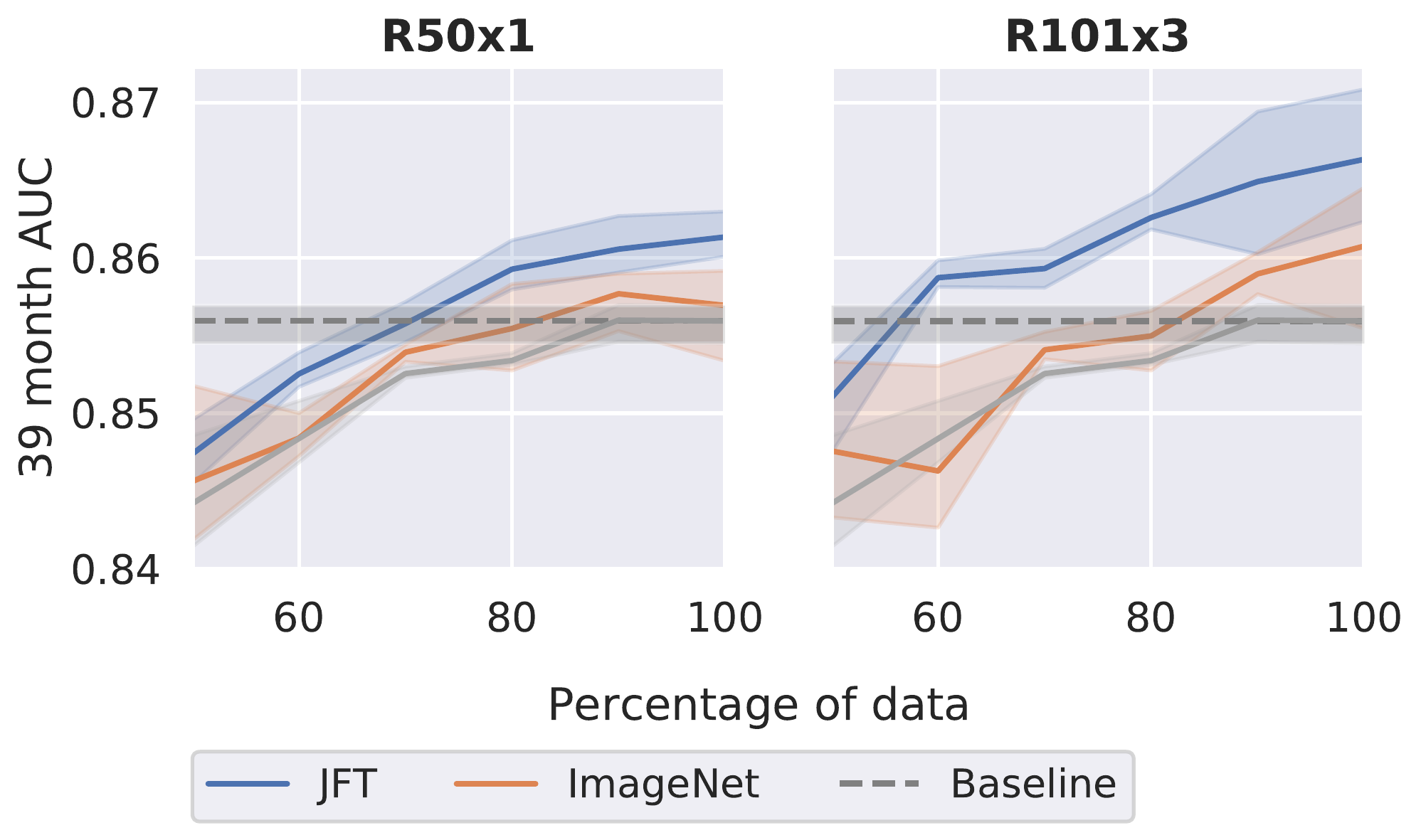}
        \caption{Impact on reduced data on 39 month AUC}
        \label{fig:mammo_data_efficiency}
     \end{subfigure}
     \begin{subfigure}[b]{0.4\textwidth}
         \centering
         \includegraphics[height=1.7in]{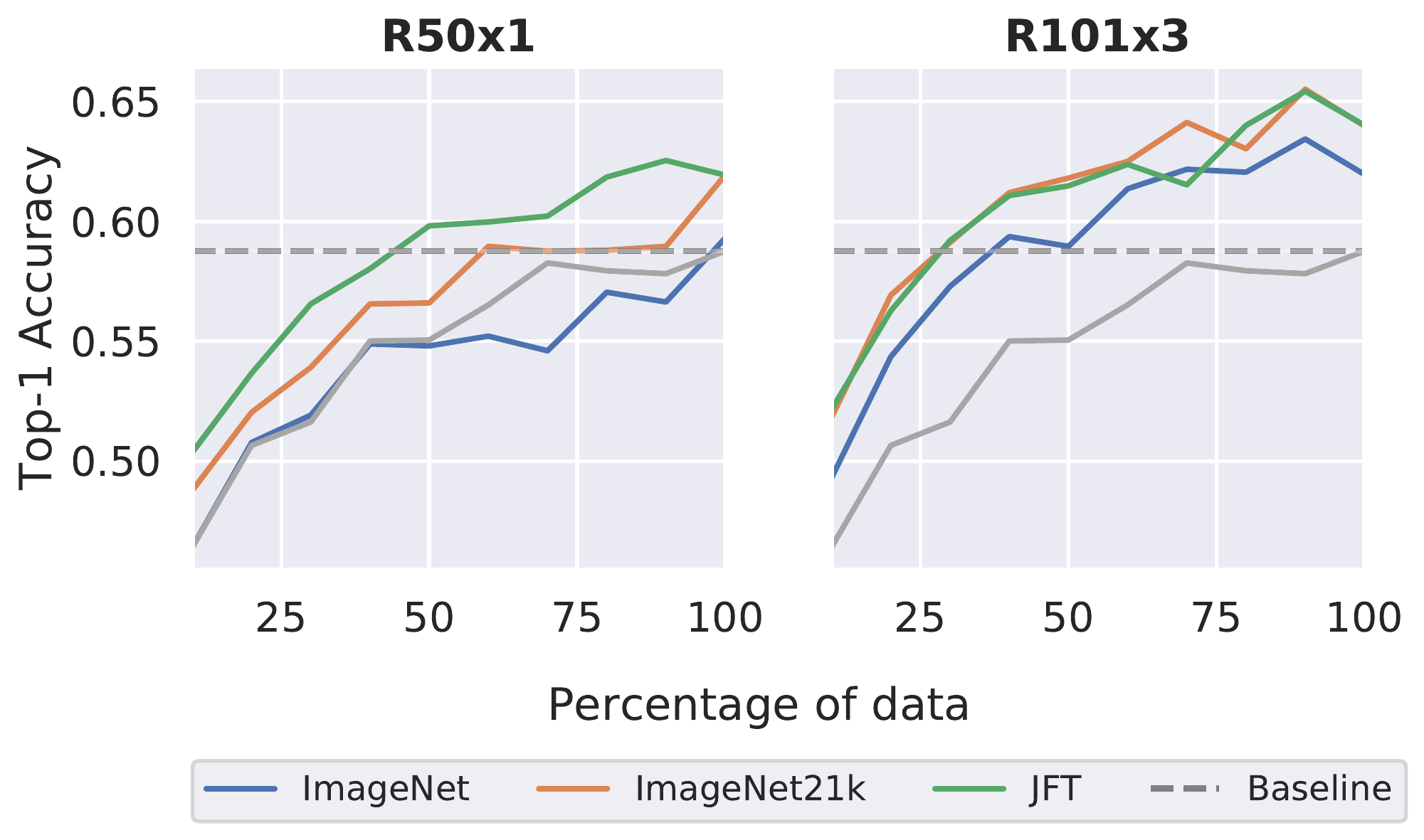}
        \caption{Impact of reduced data on dermatology accuracy}
        \label{fig:derm_data_efficiency}
     \end{subfigure}
     \begin{subfigure}[b]{\textwidth}
         \centering
         \includegraphics[height=1.7in]{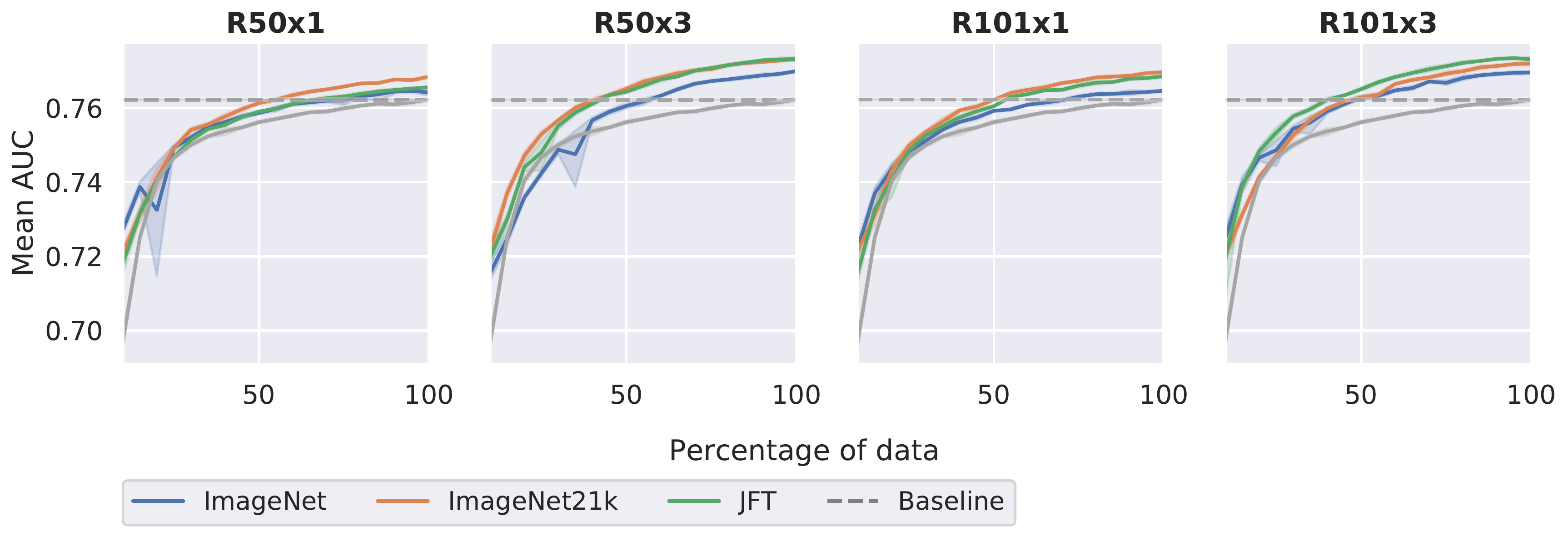}
        \caption{Impact of reduced data on CheXpert mean AUC}
        \label{fig:chexpert_data_efficiency}
     \end{subfigure}
    \caption{Data efficiency curves - improvements from large scale transfer are predominately retained at all data fractions.}
    \label{fig:data_efficiency_all}
\end{figure*}

Figures~\ref{fig:data_efficiency_all} show the decay in performance with respect to the amount of data. The BiT models retain their superiority over the baseline for all data fractions, but relative to the performance at full data, no models display a clearly different dropoff rate.

\subsection{Subgroup Analysis}
\label{app:fairness}
\begin{figure*}
    \centering
    \includegraphics[height=2.1in]{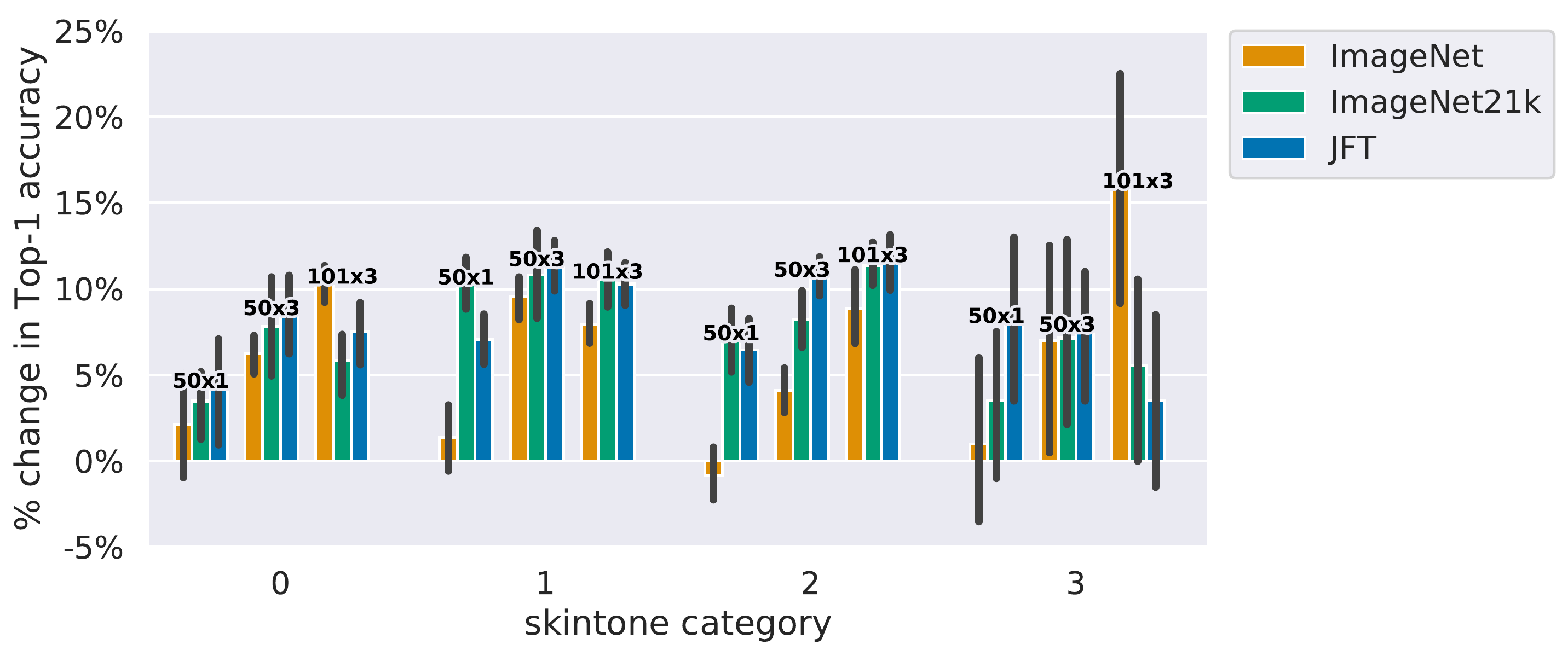}
    \caption{Dermatology subgroup performances relative to baseline}
    \label{fig:derm_subgroups}
\end{figure*}

\begin{figure*}
    \centering
    \includegraphics[height=2.1in]{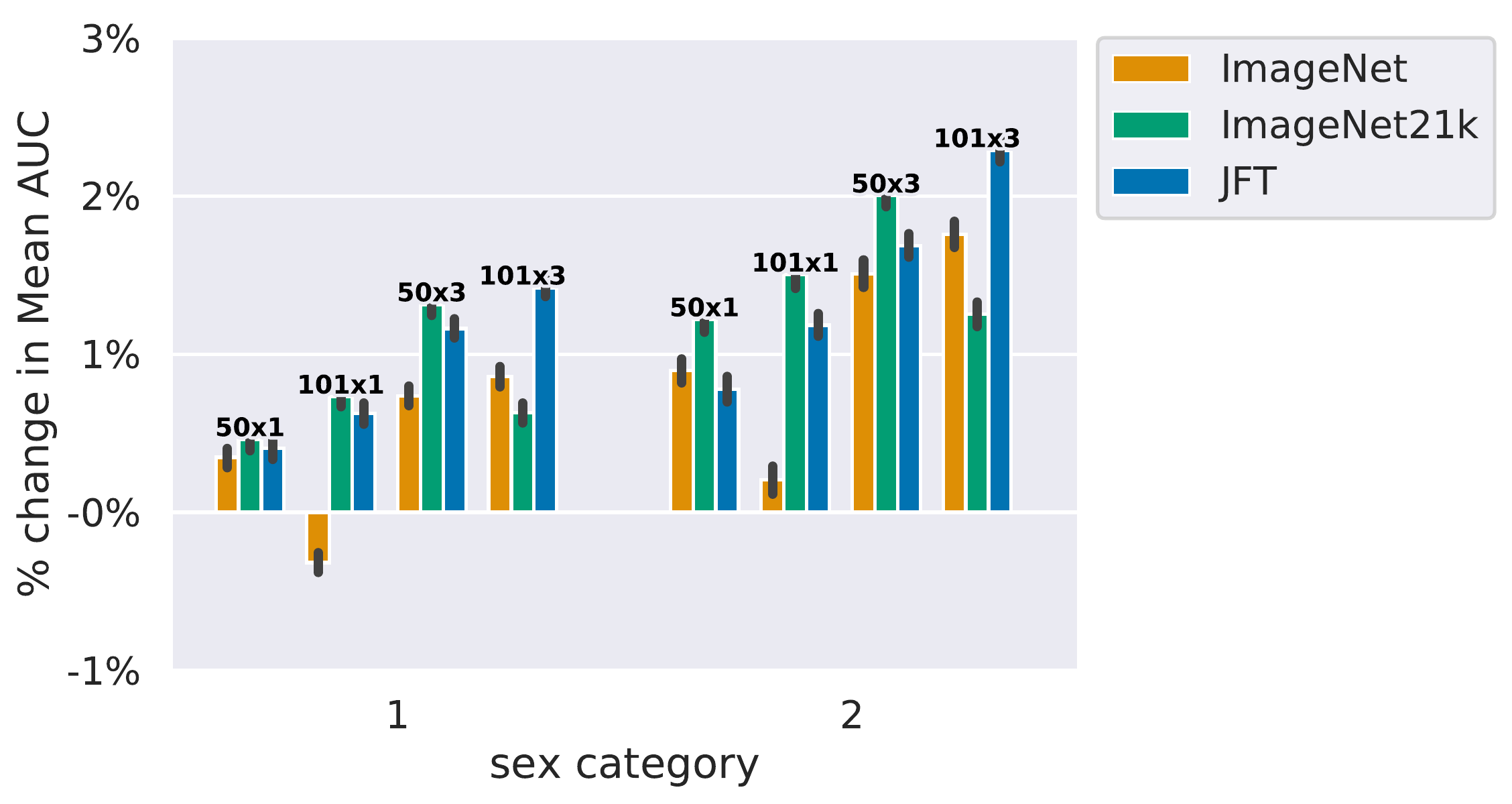}
    \caption{Chest X-Ray subgroup performances relative to baseline}
    \label{fig:chexpert_subgroups}
\end{figure*}

\begin{figure*}
    \centering
    \includegraphics[height=2.1in]{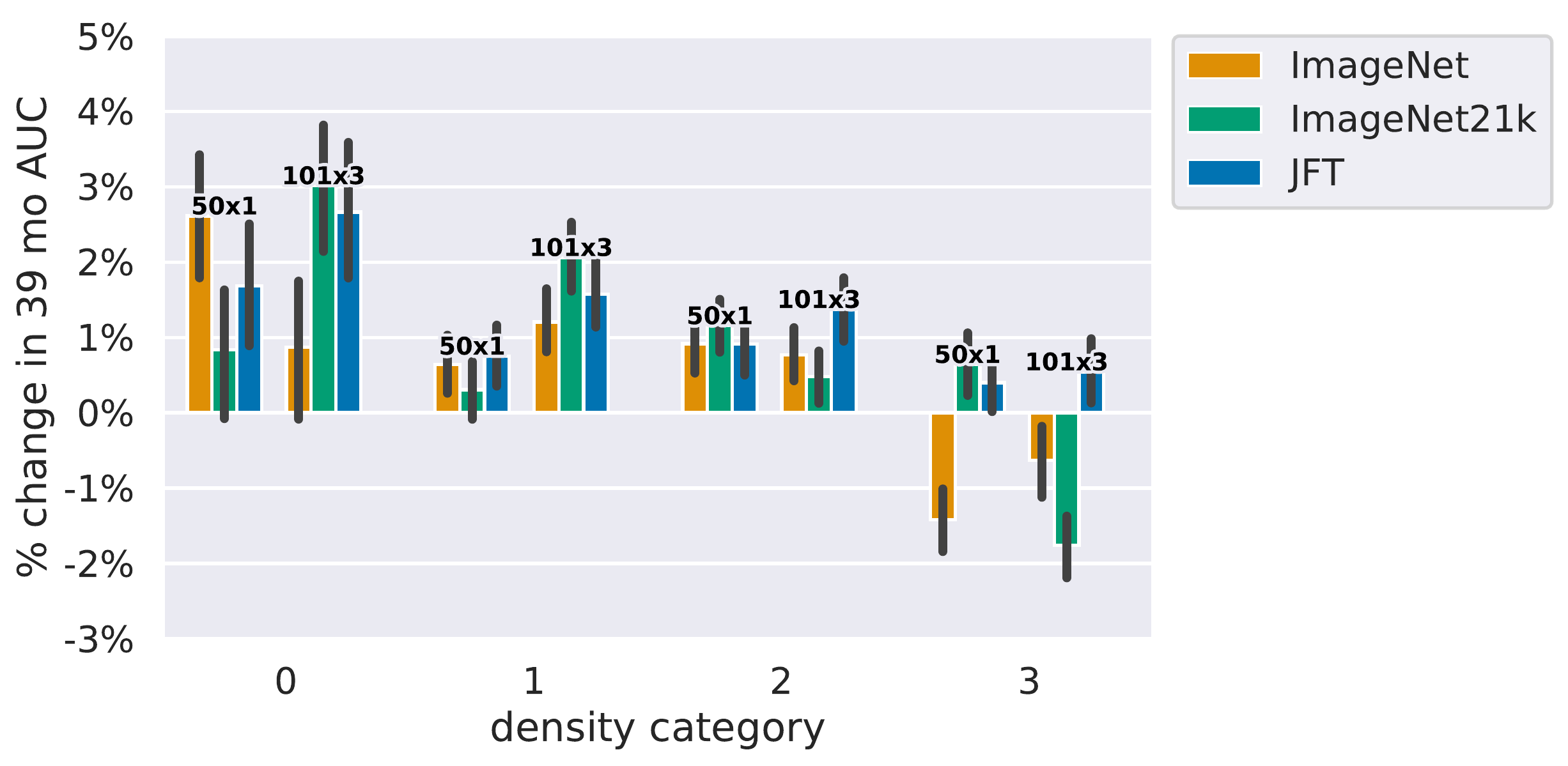}
    \caption{Mammography subgroup performances relative to baseline}
    \label{fig:mammo_subgroups}
\end{figure*}

Figures~\ref{fig:derm_subgroups},~\ref{fig:chexpert_subgroups} and~\ref{fig:mammo_subgroups} show performances relative to the baseline for different skin-tone, sex and breast density subgroups in the dermatology, chest X-Ray and mammography models respectively. The first thing we note is the lack of a particular trend. It does not seem clear that scaling pre-training data or architecture scale systematically disadvantages or advantages specific subgroups. Generally, all subgroups see some kind of performance improvement against the baseline, though the distribution of improvements across subgroups isn't always even. The one exception to this is the final mammography density category. There is a decrease in performance for this subgroup when scaling up to a R101x3, but this is ameliorated by initialisation with the JFT pre-training.\\
Note that we strongly believe that for a more complete study of these phenomena, metrics capturing performance equality across subgroups should simply be included in the model selection process. Nonetheless it is useful to see an example of the kinds of trade-offs happening behind the scenes in the setup often used by practitioners; training models to improve aggregate metrics.

\subsection{Resolution}
\begin{figure*}
    \centering
    \includegraphics[width=0.6\textwidth]{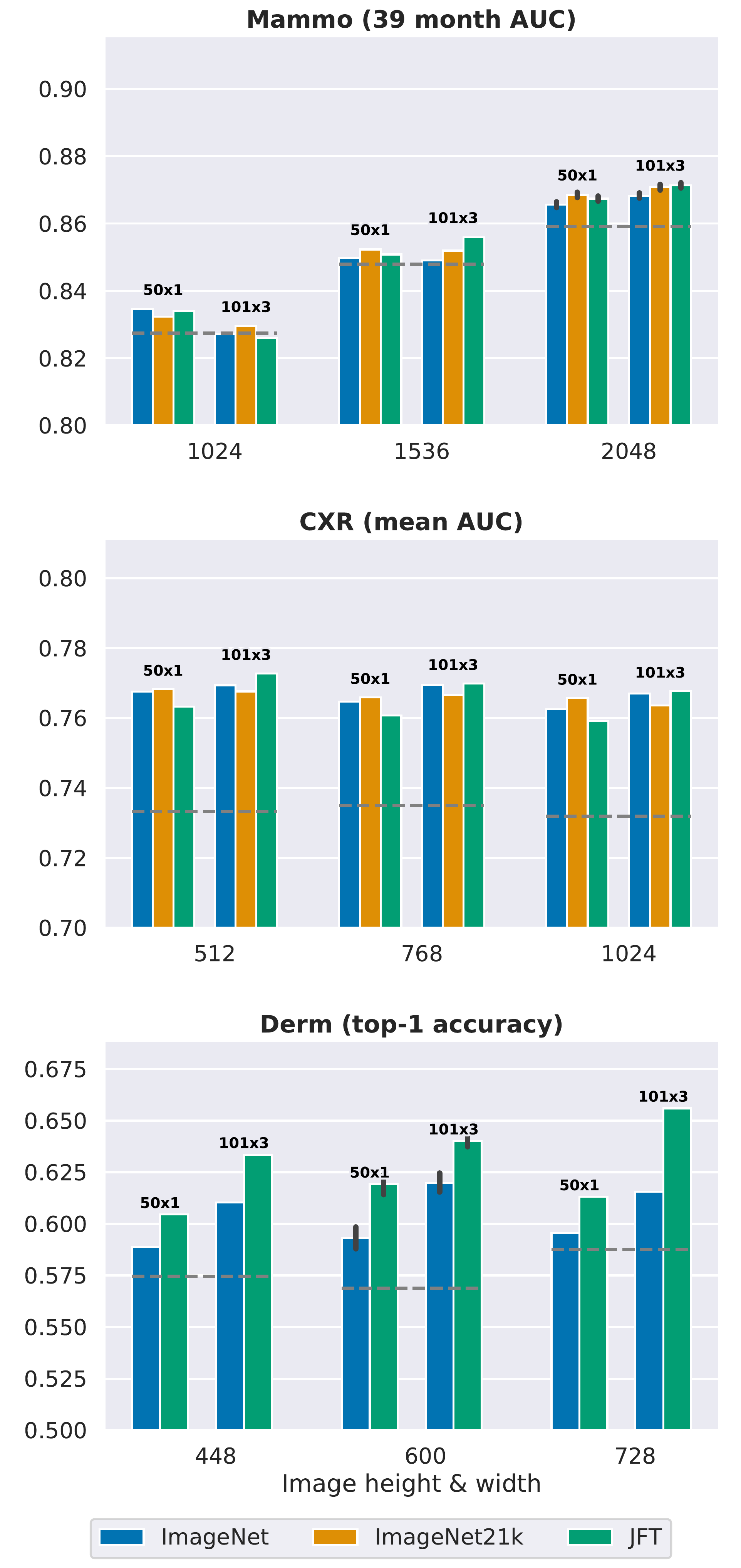}
    \caption{Absolute performance of BiT models and baseline with across resolutions. Broadly, across Mammography and Derm, models performed better at higher resolutions, and this increase was larger when using large-scale architectures and pre-training data. For Chexpert, model performance did not change with changes in image resolution. \textit{Note: all images in all tasks were square.}}
    \label{fig:resolution-result-extra}
\end{figure*}

In figure~\ref{fig:resolution-result-extra}, we demonstrate the relative change in performance of our models at different resolutions compared to baseline. here, we also show the results in absolute terms, including the finding that neither the baseline nor BiT models perform differently across different resolutions.

\subsection{Calibration}
\label{app:calibration}
\begin{figure*}[h]
     \centering
     \includegraphics[width=0.75\textwidth]{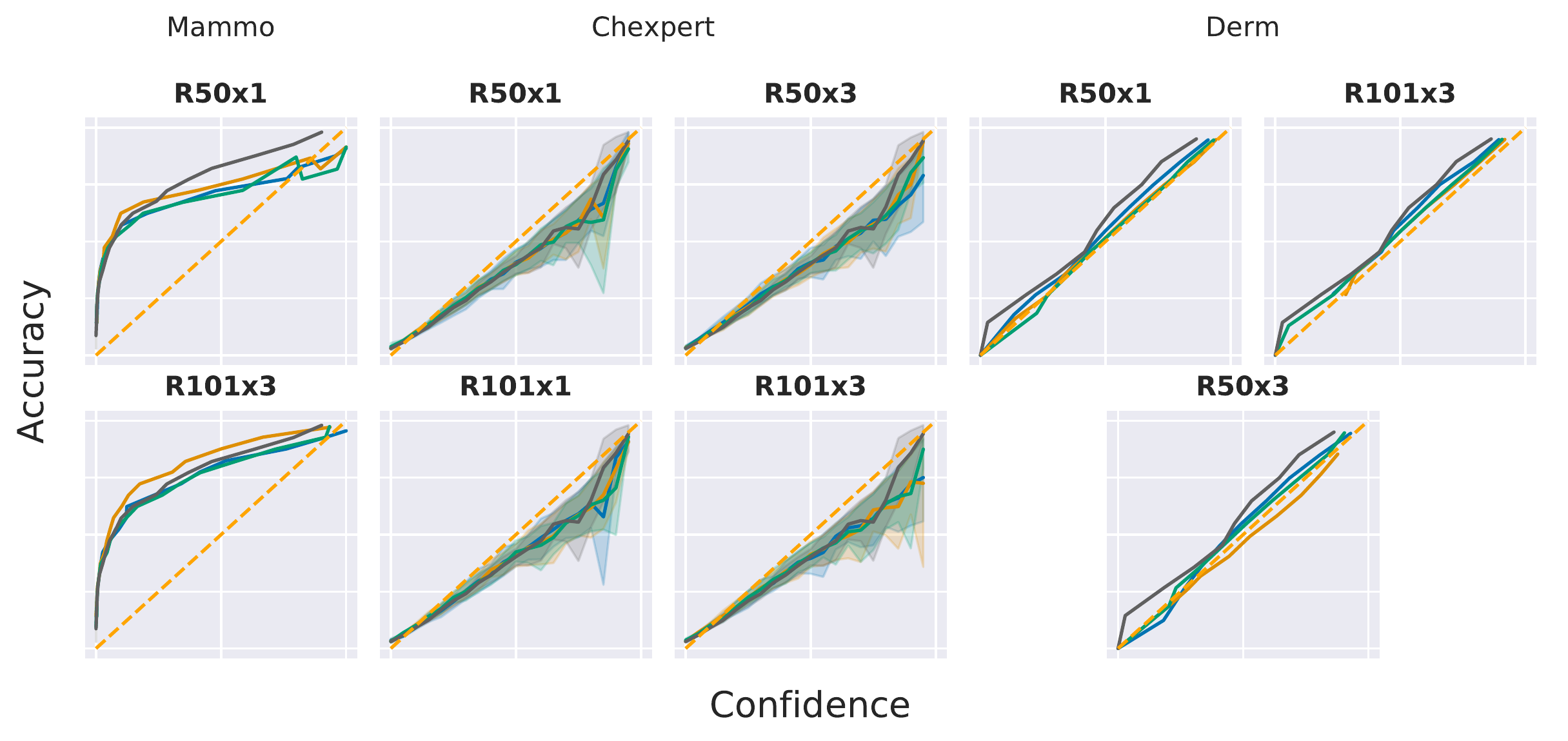}
    \caption{Reliability diagrams for different models. Across the spectrum we noted little deviation in model calibration; this was also reflected in broadly similar expected calibration errors. Shaded area for chexpert = variation across conditions.}
    \label{fig:reliability_diagrams}
\end{figure*}

Figure~\ref{fig:reliability_diagrams} shows the reliability diagrams for different models. We noted that the architecture and pre-training wasn't the controlling factor for the calibration levels of these models.

\subsection{Weight \& feature analysis}
\label{app:weight_feature_analysis}
\begin{figure*}[ht]
     \centering
     \begin{subfigure}[b]{1.0\textwidth}
         \centering
         \includegraphics[width=\textwidth]{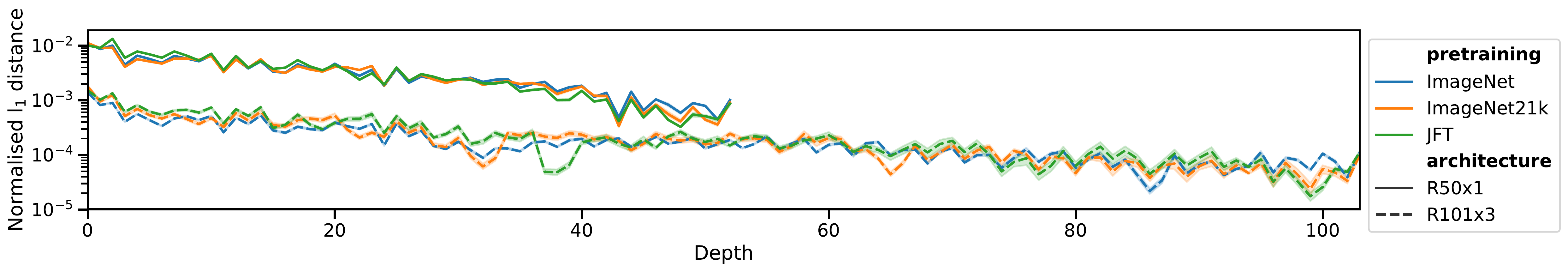}
        \caption{Mammography models - 5 trials.}
        \label{fig:all_mammo_weights}
     \end{subfigure}
     \begin{subfigure}[b]{1.0\textwidth}
         \centering
         \includegraphics[width=\textwidth]{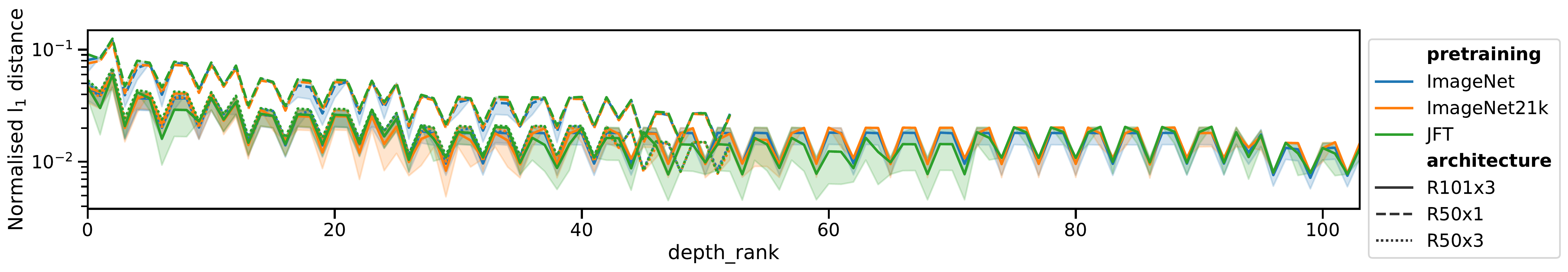}
        \caption{Dermatology models - 10 trials. \textit{R50x3 largely overlaps with the R101x3, indicating the dynamics may be more controlled by width than depth.}}
        \label{fig:all_derm_weights}
     \end{subfigure}
    \caption{l\textsubscript{1} distance between final trained model kernels and the initialisation, normalised by kernel size. Larger models move less in parameter space. Shaded area = 95\% bootstrapped confidence intervals.}
    \label{fig:all_weights}
\end{figure*}

Here we show full figures for the weight-space analysis. Figure~\ref{fig:all_weights} shows the deviation in parameter space of models from their intial checkpoint. Note we expect Mammo and Derm model weights not to be directly comparable due to the use of different optimizers/regularisers,
\section{Qualitative examples}
\label{app:examples}
To confirm that performance improvements from BiT checkpoints do not originate from spurious correlations between image features and outcomes, which larger models pretrained on diverse data could more effectively exploit, we performed a brief qualitative analysis. We specifically highlight three images in which BiT 50x1 models  correctly predict the development of breast cancer while the baseline model does not.

\begin{figure*}
    \centering
    \includegraphics[width=0.9\textwidth]{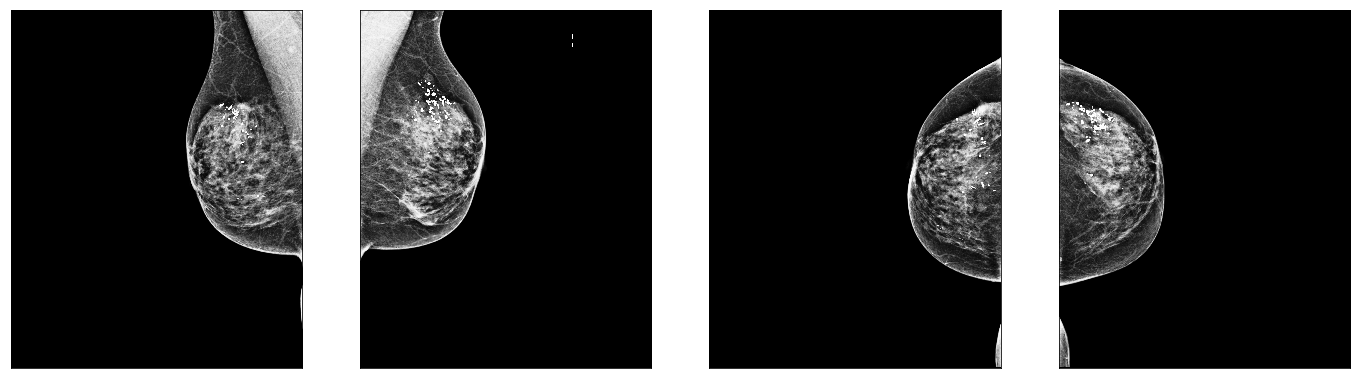}
    \caption{An example of a cancer detection on which a BiT model outperforms the baseline model. This image contains microcalcification groupings, a finding that may indicate increased cancer risk \cite{ali2019association}. This is an indication that BiT models pick up on real cancer risk factors that the baseline model may miss. \textbf{Ground truth}: cancer developed within 39 months. \textbf{Baseline score}: 0.48; \textbf{BiT-L 50x1 score}: 0.72.}
    \label{fig:mammo_qual_mcc}
\end{figure*}

This analysis showed no spurious artifacts or correlations with the data (see e.g. \ref{fig:mammo_simple_example}. And, while the actual identification of how and why a cancer develops is very hard to know, it also showed that the BiT models outperformed on mammograms that showed microcalcification groups (see e.g. \ref{fig:mammo_qual_mcc}, which can be associated with certain cancer developments and are used both by humans and computer-aided diagnostic programs to identify high-risk mammograms \cite{ali2019association}.

\begin{figure*}
    \centering
    \includegraphics[width=0.9\textwidth]{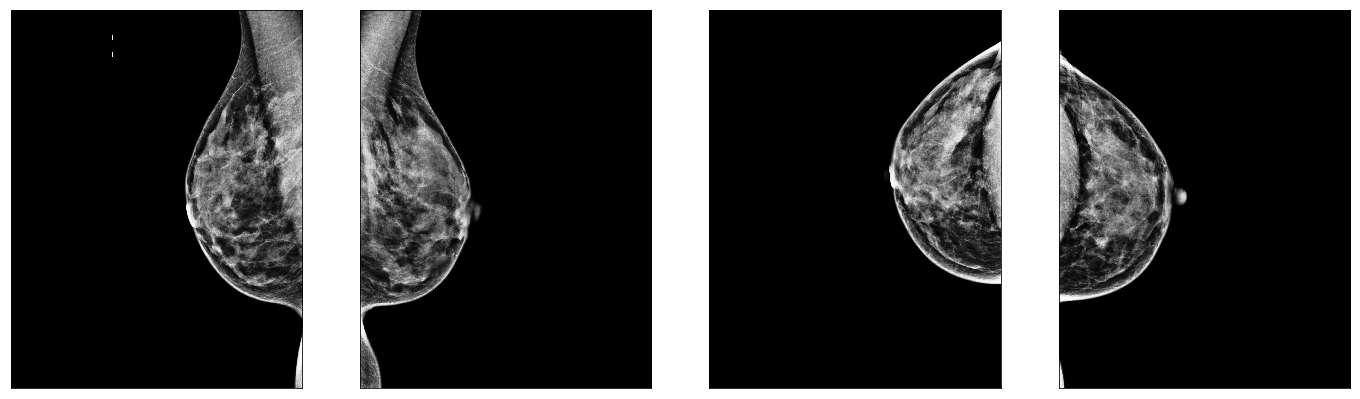}
    \caption{An example of a cancer detection on which a BiT model outperforms the baseline. There are no artifacts or spurious effects in this image. \textbf{Ground truth}: cancer developed within 39 months. \textbf{Baseline score:} 0.42; \textbf{BiT-L 50x1 score:} 0.70.}
    \label{fig:mammo_simple_example}
\end{figure*}
\begin{figure*}
    \centering
    \includegraphics[width=0.9\textwidth]{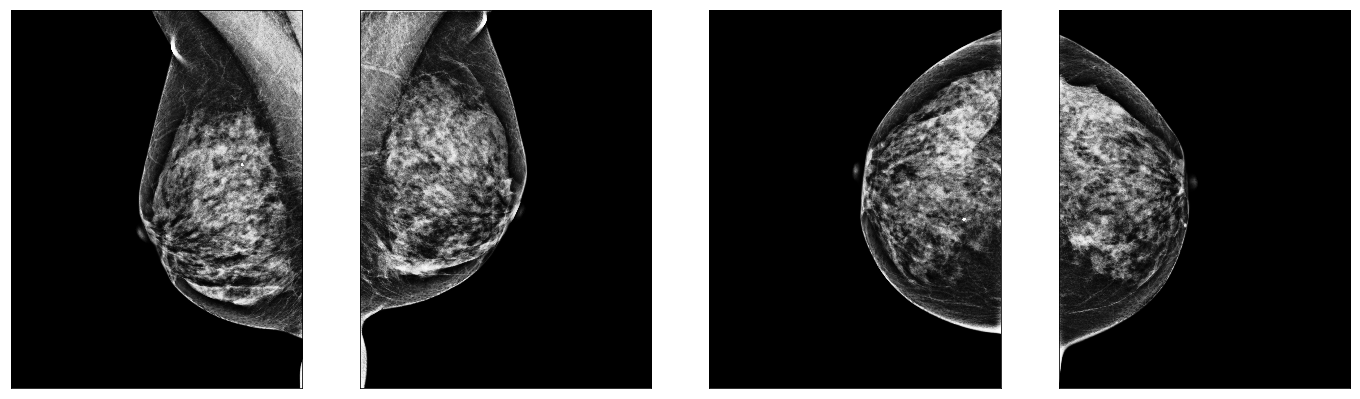}
    \caption{An example of a cancer detection on which a BiT model outperforms the baseline model. This mammogram depicts high-density breasts, which both human experts and machine learning models tend to perform worse on \cite{mckinney2020international}. \textbf{Ground truth}: cancer developed within 39 months. \textbf{Baseline score}: 0.32; \textbf{BiT-L 50x1 score}: 0.62.}
    \label{fig:mammo_qual_density}
\end{figure*}
\begin{figure*}
    \centering
    \includegraphics[width=0.5\textwidth]{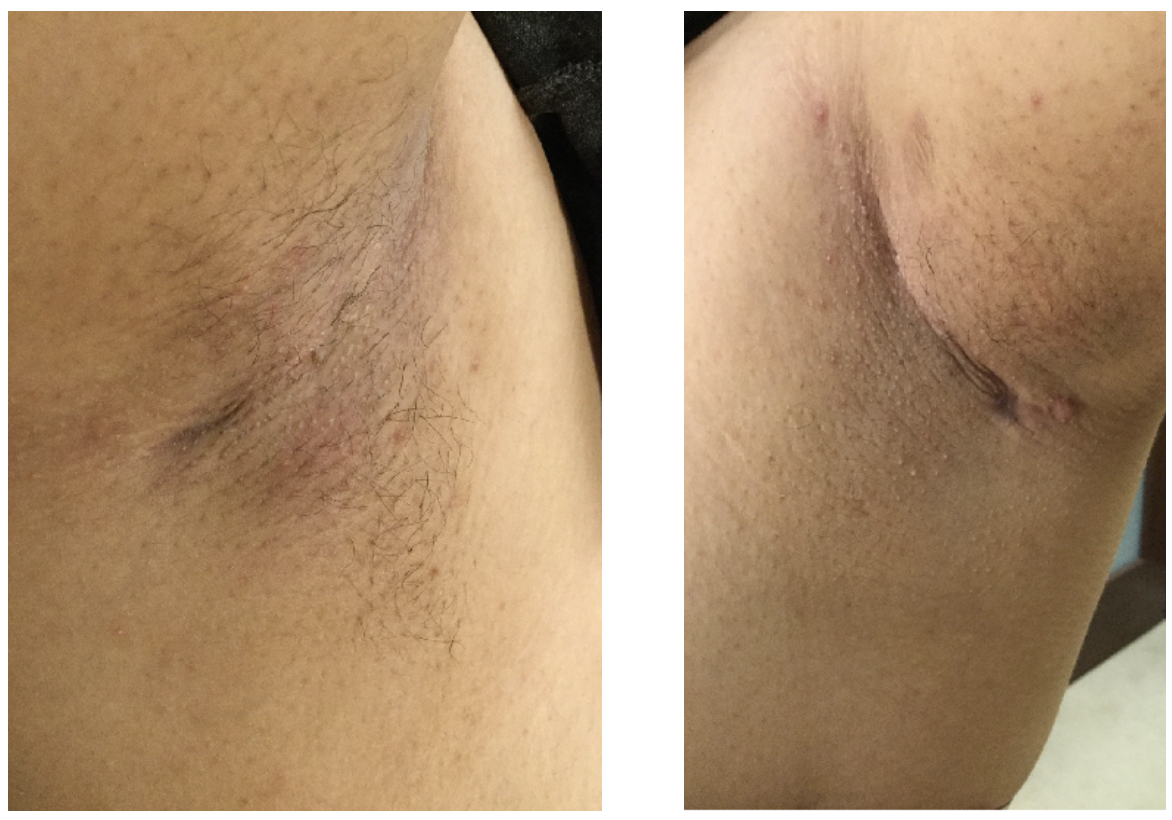}
    \caption{An example of a dermatology case on which a BiT model outperforms the baseline. The image contains very little background, evidence that the model is utilizing information solely from the skin. \textbf{Ground truth}: Hidradenitis. \textbf{Baseline score:} 0.33; \textbf{BiT-L 50x1 score:} 0.91.}
    \label{fig:mammo_simple_example}
\end{figure*}

\end{document}